\documentclass[journal]{IEEEtran}
\ifCLASSINFOpdf
\else
\fi
\usepackage{url}
\usepackage{booktabs}
\usepackage{multirow}

\usepackage{mathrsfs}
\usepackage{verbatim}
\usepackage{amsmath,amssymb,amsfonts}
\usepackage{lineno}
\usepackage{stfloats}
\usepackage{float}
\usepackage{graphicx}
\usepackage{subfigure}
\usepackage{epstopdf}
\usepackage{textcomp}
\usepackage{bm}

\usepackage{hyperref}
\usepackage[linesnumbered,ruled,lined,boxed]{algorithm2e}

\usepackage{adjustbox}
\usepackage[numbers,sort&compress]{natbib}
\usepackage{threeparttable}

\usepackage{makecell}

\usepackage{comment}

\usepackage{color} 
\begin{document}
\newcommand{\etal}{\textit{et al}. }
\newcommand{\ie}{\textit{i}.\textit{e}., }
\newcommand{\eg}{\textit{e}.\textit{g}. }
\renewcommand{\figureautorefname}{Fig.}       
\renewcommand{\tableautorefname}{Tab.}        
\renewcommand{\equationautorefname}{Eq.}      
\renewcommand{\sectionautorefname}{Sec.}      
\renewcommand{\subsectionautorefname}{Sec.}   
\renewcommand{\subsubsectionautorefname}{Sec.}
\renewcommand{\appendixautorefname}{App.}     

\title{Class Incremental Fault Diagnosis under Limited Fault Data via Supervised Contrastive Knowledge Distillation}
%
%
%

\author{Hanrong Zhang$^\ast$, Yifei Yao$^\ast$, Zixuan Wang$^\ast$, Jiayuan Su, Mengxuan Li, Peng Peng, Hongwei Wang$^\dagger$
\thanks{
Hanrong Zhang, Jiayuan Su, Peng Peng, and Hongwei Wang are with ZJU-UIUC Joint Institute at Zhejiang University, Haining 314400, China (E-mail: zhanghr0709@gmail.com, \{jiayuan.23, pengpeng, hongweiwang\}@intl.zju.edu.cn). Yifei Yao is with the ZJU-UoE Institute at Zhejiang University, Haining 314400, China (E-mail: yifei3.23@intl.zju.edu.cn). Zixuan Wang is with the College of Biomedical Engineering and Instrument Science at Zhejiang University, Hangzhou 310013, China (E-mail: zixuanw.20@intl.zju.edu.cn). Mengxuan Li is with the College of Computer Science and Technology in Zhejiang University Hangzhou, 310013, China (E-mail: mengxuanli@intl.zju.edu.cn).
}%
}
\markboth{IEEE Transactions on Industrial Informatics}%
{Shell \MakeLowercase{\textit{et al.}}: Enhance Fault Diagnosis Performance By Neural Network Ari}
%



\maketitle

\begin{abstract}

Class-incremental fault diagnosis requires a model to adapt to new fault classes while retaining previous knowledge. However, limited research exists for imbalanced and long-tailed data. Extracting discriminative features from few-shot fault data is challenging, and adding new fault classes often demands costly model retraining. Moreover, incremental training of existing methods risks catastrophic forgetting, and severe class imbalance can bias the model's decisions toward normal classes.
To tackle these issues, we introduce a \textbf{S}upervised \textbf{C}ontrastive knowledge disti\textbf{L}lation for class \textbf{I}ncremental \textbf{F}ault \textbf{D}iagnosis (SCLIFD) framework proposing supervised contrastive knowledge distillation for improved representation learning capability and less forgetting, a novel prioritized exemplar selection method for sample replay to alleviate catastrophic forgetting, and the Random Forest Classifier to address the class imbalance. Extensive experimentation on simulated and real-world industrial datasets across various imbalance ratios demonstrates the superiority of SCLIFD over existing approaches. Our code can be found at \url{https://github.com/Zhang-Henry/SCLIFD_TII}.
\end{abstract}


%
\IEEEpeerreviewmaketitle

\section{Introduction}
\label{intro}
Data-driven fault diagnosis techniques have gained significant prominence over the past two decades \cite{10106026, 9780560, li2023order, wang2023hard,10510599}. However, most of them necessitate sufficient training data to achieve reliable modeling performance\cite{10114639,10152774,li2023sccam,wang2023imbalanced}.
Unfortunately, fault data is typically limited in comparison to normal data. This is because engineering equipment primarily operates under normal conditions, and the probabilities of faults vary across different working environments. Besides, fault simulation experiments are costly and inevitably deviate to some extent from real industrial environments. These possible reasons consequently contribute to class imbalance and a long-tailed distribution among different conditions \cite{Chen_Chen_Feng_Liu_Zhang_Zhang_Xiao_2022}. The performance of the model typically suffers as it tends to prioritize the normal class, consequently neglecting fault classes or tail classes. Therefore, extensive research efforts have been devoted to addressing this challenge, leading to significant advancements in the field. 

Class-imbalanced and long-tailed fault diagnosis are usually effectively addressed by 
data resampling, cost-sensitive learning, and information augmentation~\cite{Chen_Chen_Feng_Liu_Zhang_Zhang_Xiao_2022}. Concretely, data resampling mitigates the class imbalance issue by reconstructing a class-balanced dataset \cite{Khoshgoftaar_Gao_2009a}. 
Cost-sensitive learning aims to compensate for the impact of class imbalance during the training process by adjusting the weights of different classes' influence on the target function \cite{Liu_Li_Zio_2017}. 
Information augmentation-based approaches, which primarily involve data generation and transfer learning, aim to address the imbalance problem by leveraging auxiliary data or diagnostic experience from other datasets. Specifically, data generation utilizes generative models such as generative adversarial networks \cite{Goodfellow_Pouget-Abadie_Mirza_Xu_Warde-Farley_Ozair_Courville_Bengio_2020} and variational auto-encoders \cite{Li_Jiang_Liu_Zhang_Xu_2021}, to produce additional samples, thereby augmenting the dataset by simulating the original distribution of the source data. 
Transfer learning seeks to enhance model performance by transferring information, such as data and features, from a source domain to a target domain \cite{Zhiyi_Haidong_Lin_Junsheng_Yu_2020}.
\cite{chu2020feature} use class activation maps to separate features into class-specific features and common ones, which are then combined to augment tail classes.
 

Besides the challenge posed by the limited quantity and variety of fault samples available for real-world industrial fault diagnosis, additional fault samples from new classes, \textit{i}.\textit{e}., incremental samples of new fault classes, can be continuously obtained as the industrial process progresses. However, the methods mentioned above only concentrate on mitigating the class imbalance issue but totally overlook the model's adaptation ability to recognize fault samples from new classes encountered by fault diagnosis models. As a result, when employing the fault diagnosis methods mentioned above in incremental scenarios, the need to abandon prior efforts and retrain the well-trained model each time new class samples emerge incurs a significant computational cost and time expense. Furthermore, as incremental learning progresses and the model is trained on new types of data, it may encounter \textit{catastrophic forgetting} \cite{McCloskey_Cohen_1989,yao2024uncertaintyclarityuncertaintyguidedclassincremental}, \textit{i}.\textit{e}., the new learned knowledge may overwrite the old previously learned knowledge, so the model performance continually declines. Therefore, effective class incremental fault diagnosis methods are also essential to mitigate catastrophic forgetting.

Currently, there is growing interest in incremental learning methods for fault diagnosis\cite{peng2023sclifdsupervisedcontrastiveknowledgedistillation}. \cite{Shi_Ding_Chang_Shen_Huang_Zhu_2024} propose a CDCIBN framework for cross-domain incremental fault diagnosis. \cite{Hell_Pestana_Soares_Goliatt_2022} introduce a Self-Organizing Fuzzy classifier to handle large-scale, streaming data in dynamic environments. \cite{Zheng_Xiong_Zhang_Su_Hu_2022} use Repeated Replay with Memory Indexing for bearing defect diagnostics, retaining knowledge across varied settings. \cite{Yu_Zhao_2020} suggest a broad convolutional neural network that incrementally learns new fault samples. \cite{Gu_Zhao_Yang_Li_2022} introduce an incremental imbalance-modified CNN for chemical fault diagnosis. \cite{Ren_Liu_Wang_Zhang_2022} propose HSELL-Net, leveraging lifelong learning and domain adaptation for small sample enhancement. However, existing methods often overlook real-world few-shot scenarios or require training with fault samples from other conditions \cite{shaoIFDCNN,zhangOpenDA,liuDeepSunDA,liangVariable,Zhang2024MultiscaleCA}.

\begin{figure}[t]
    \centering
    \includegraphics[width=0.4\textwidth]{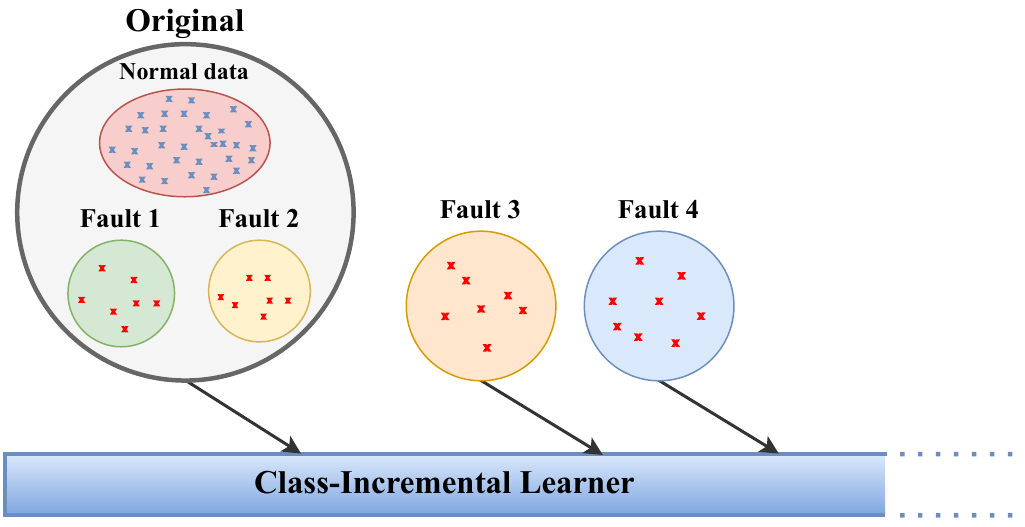}
    \caption{Class incremental learning for fault diagnosis under limited fault data.}
    \label{incremental}
\end{figure}

In summary, it is necessary to propose a method to address class incremental fault diagnosis under the challenges of class-imbalanced data and long-tail distributed data, as illustrated in \autoref{incremental}. This research tackles class incremental learning in fault diagnosis, addressing class imbalance and long-tailed distributions common in the industry. Our framework handles limited fault data, enhancing diagnostic accuracy and reliability, and bridging the gap between theory and practical industrial applications for complex system maintenance.
Our contributions are summarized as follows:
\begin{enumerate}
    \item We propose an effective fault diagnosis framework SCLIFD (\textbf{S}upervised \textbf{C}ontrastive Knowledge Disti\textbf{L}lation for Class \textbf{I}ncremental \textbf{F}ault \textbf{D}iagnosis under Limited Fault Data), to creatively address the issues existing in the class incremental learning under the class imbalanced and long-tailed fault diagnosis. 
    \item We propose a novel Supervised Contrastive Knowledge Distillation (SCKD) to address the challenges of learning discriminative features from limited fault samples and mitigating catastrophic forgetting. Unlike Supervised Contrastive Learning~\cite{Khosla_Teterwak_Wang_Sarna_Tian_Isola_Maschinot_Liu_Krishnan_2020}, which enhances feature representation within a single session, our method introduces cross-session knowledge transfer. SCKD distills knowledge from the previous session’s feature extractor to the current session’s, ensuring the model retains past knowledge while learning new fault classes. Unlike typical knowledge distillation that focuses on logits, SCKD distills feature representations, preserving discriminative features across both old and new classes.

    \item We propose a novel prioritized exemplar selection method Marginal Exemplar Selection (MES) to mitigate catastrophic forgetting by sample replay. It reserves the most challenging exemplars of each class that reside on the peripheries of each class's feature space into a memory buffer for subsequent incremental training.
    \item We adopt the Balanced Random Forest Classifier~\cite{Chen2004UsingRF} for classification due to its proficiency in equalizing class distribution and reducing classification bias through ensemble learning. This approach improves the recognition of minority fault classes and enhances overall diagnostic performance in imbalanced data scenarios.
    \item We evaluate the effectiveness of the proposed fault diagnosis method using simulated Tennessee Eastman Process (TEP) and practical Multiphase Flow Facility (MFF) datasets, both characterized by imbalanced and long-tailed distributions. The results demonstrate that our model outperforms other state-of-the-art methods, achieving the best performance.
\end{enumerate}

The rest of the paper is organized as follows. In  \autoref{background}, the observations and motivations are highlighted, and the preliminary theories of this work are introduced. The details of the proposed method are delved into in  \autoref{method}. The comparative and ablation experiments performed on the TEP and MFF are explained in  \autoref{Experiment}. Finally, \autoref{conclusion} concludes this paper.

\section{Motivations and Background Theory}
\label{background}

\subsection{Observations and Motivations}


\begin{figure}[ht]
  \centering
  \begin{minipage}[b]{0.43\linewidth}
    \includegraphics[width=\linewidth]{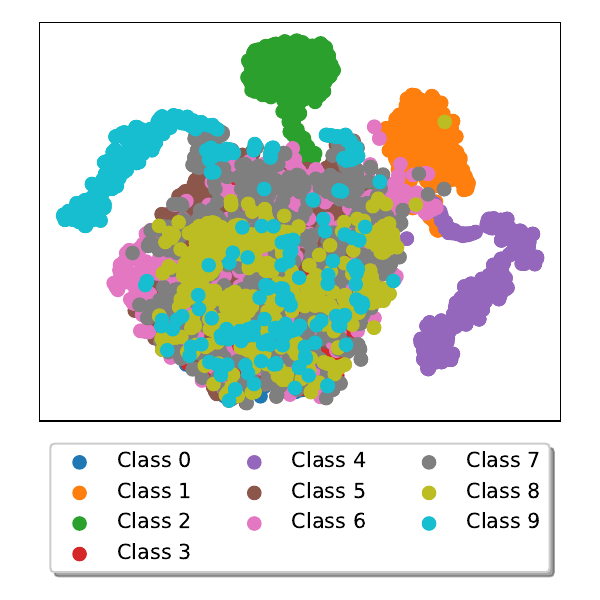}
    \caption{iCaRL's T-SNE visualization under the imbalanced case of TEP dataset. iCaRL cannot extract discriminative features from imbalanced classes.}
    \label{fig:ob-icarl}
  \end{minipage}
  \quad 
  \begin{minipage}[b]{0.5\linewidth}
    \includegraphics[width=\linewidth]{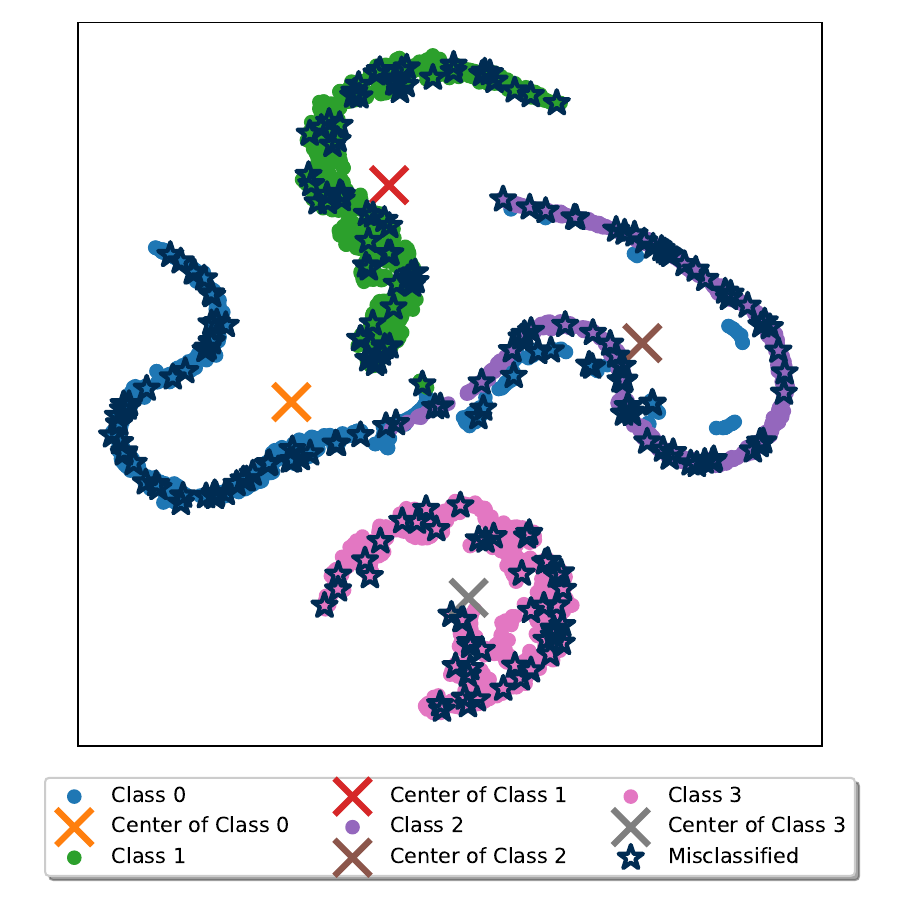}
    \caption{T-SNE visualization under the long-tailed case of MFF dataset using ``Herding''~\cite{icarl,welling2009herding}. Misclassified samples tend to be scattered across the model's decision boundaries.}
    \label{fig:ob-sclifd}
  \end{minipage}
\end{figure}

As we described in \autoref{intro}, it is vital to address the imbalanced and long-tailed fault diagnosis in the class incremental learning setting, but few works tackle the circumstances. Moreover, as shown in \autoref{tab:baseline} in our experiments, we find traditional frameworks for class incremental learning in the balanced class setting, such as iCaRL~\cite{icarl}, EEIL~\cite{Castro_2018_ECCV} also have poor performances under imbalanced and long-tailed settings.  To find out the reasons behind their poor performances, we have the following experiments and observations.

Initially, we find that the representation learning capability of iCaRL~\cite{icarl}, which utilizes cross-entropy loss as the classification loss, is insufficient to extract discriminative features from limited fault data. 
We visualize the T-SNE feature embeddings extracted by iCaRL~\cite{icarl} under the imbalanced case of the TEP dataset. We have observed that in the T-SNE visualization, features of different classes are intertwined and cannot be clearly distinguished. 
Therefore, we consider adopting a powerful representation learning method Supervised Contrastive Learning~\cite{Khosla_Teterwak_Wang_Sarna_Tian_Isola_Maschinot_Liu_Krishnan_2020} to boost its representation learning performance. Experiments prove that SCL combined with knowledge distillation effectively enhances the performance of our method.

Furthermore, we observe that the sample replay method called ``Herding''~\cite{welling2009herding} employed in iCaRL~\cite{icarl} is not that effective in our imbalanced and long-tailed scenarios. ``Herding'' is used to mitigate catastrophic forgetting~\cite{icarl}. It uses a memory buffer to keep a subset of previous data closest to the centers of overall distributions of previous classes and mixes it into the training process as new class data is introduced. As shown in \autoref{fig:ob-sclifd}, we adapt the Herding technique in our method under long-tail fault diagnosis using the MFF dataset. By employing T-SNE visualization, we have illustrated the feature distribution and centroids for each class. 
We observe that samples misclassified by the model are often not the ones nearest to the class feature centroids. Contrarily, these samples tend to be scattered across the model's decision boundaries, lying far from the centroids. This insight led us to propose the Marginal Exemplar Selection (MES) method, which aims to identify and select samples that are located at the edges of the feature space and lie on the boundaries of the data distribution. This could enhance the model's boundary recognition and generalization capabilities by focusing on marginal samples hard to distinguish.

Finally, it is challenging to make fault classification under limited fault data, because the model is inclined to focus on the normal class of a large amount while neglecting the fault classes of few shot numbers~\cite{Li_Peng_Zhang_Wang_Shen_2024}.
Consequently, the emphasis of the model training is biased toward the normal class, hence giving rise to classification favoring the normal class. Therefore, to handle the problem, we adopt the BRF classifier due to its ability to even out class distribution, utilize ensemble learning to diminish bias and improve the model's capability to differentiate among all classes accurately. This ensures equitable and precise classification across both old and new classes.

\begin{figure*}[t]
    \centering
    \includegraphics[width=\textwidth]{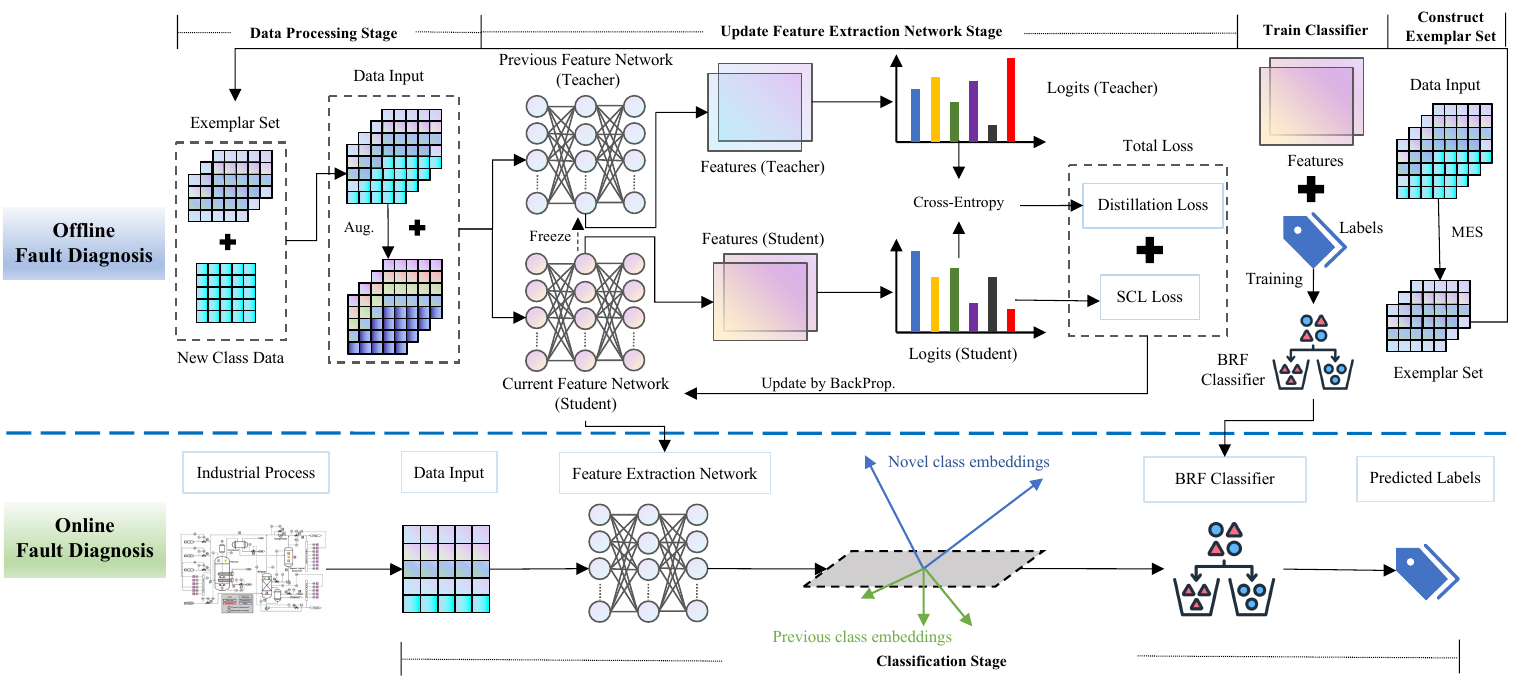}
    \caption{The general framework of the proposed method SCLIFD for fault diagnosis under limited fault data, which is detailed explained in \autoref{overview}.}
    \label{update}
\end{figure*}
\begin{algorithm}[t]
    
    \caption{SCLIFD Fault Diagnosis in an Incremental Session}\label{alg:train}
    \SetKwFunction{isOddNumber}{isOddNumber}

    \SetKwInOut{KwIn}{Input}
    \SetKwInOut{KwOut}{Output}
    \SetKwInOut{KwReq}{Require}
    \KwIn{Training samples in per-class sets $X^s, \ldots, X^t$; Memory size $K$; Samples to be classified $X$; Classifier $\mathcal{C}$}
    \KwReq{Current feature extractor $\varphi_{\theta}: \mathcal{X} \rightarrow \mathbb{R}^{d}$; Current memory buffer $\mathcal{P}=\left(P_1, \ldots, P_{s-1}\right)$, $\mathcal{P}_{y}=\left(p_1, \ldots, p_{|\mathcal{P}_{y}|}\right), y=1,\ldots,s-1$}

    ${\varphi_{\theta}}^{'} \leftarrow$ Update feature extractor using \autoref{alg:Feature Extractor} ($X^s, \ldots, X^t$; $\mathcal{P}$; $\varphi_{\theta}$; $\mathcal{C}$)
    
    ${\mathcal{C}}^{'}, Y^{*} \leftarrow$ Train classifier and make classification using \autoref{alg:Classification} ($\mathcal{P}$; $X$; $\varphi_{\theta}$; $\mathcal{C}$)

    $m \leftarrow K / t$   // Exemplar number per class (upper round)
    
    \For{$y=1\ldots s-1$}{
    $P_y \leftarrow\left(p_1, \ldots, p_m\right)$ // Reduce current exemplar sets
    } 

     // Prioritized new class exemplar selection using MES
    
    \For{$y=s\ldots t$}{
    $\mu_y \leftarrow $ Compute class mean using \autoref{eq:mean}
    
        \For{$k=1\ldots m$}{
        $p_k \leftarrow $ Select exemplars using \autoref{argmax}
        } 
        $\mathcal{P}_{y} \leftarrow\left(p_1, \ldots, p_m\right)$ 
    }

    $\mathcal{P}^{'}=\left(P_1, \ldots, P_{s-1}, P_{s}, \ldots, P_{t}\right)$ // Construct a new memory buffer of old and new class exemplars
    
    \KwOut{Updated feature extractor ${\varphi_{\theta}}^{'}$; New memory buffer of exemplar sets $\mathcal{P}^{'}=\left(P_1, \ldots, P_{t}\right)$; Predicted classifications  $Y^{*}$}
\end{algorithm}
\subsection{Self-Supervised Contrastive Learning and SCL}
\label{SCL}

The general concept of self-supervised contrastive learning aims to minimize the distance between each anchor and its positive sample while maximizing the distance between the anchor and other negative samples in the embedding space \cite{Khosla_Teterwak_Wang_Sarna_Tian_Isola_Maschinot_Liu_Krishnan_2020}. Specifically, let a set of $N$ randomly sampled sample/label pairs be denoted as $\left\{\boldsymbol{x}_{k}, \boldsymbol{y}_{k}\right\}_{k=1 \ldots N}$. An augmented batch comprising $2N$ pairs $\left\{\tilde{\boldsymbol{x}}_{i}, \tilde{\boldsymbol{y}}_{i}\right\}_{i=1 \ldots 2 N}$ of the same label as the source sample is generated and utilized for training. The two random augmentations of $\boldsymbol{x}_{k}(k=1 \ldots N)$ are denoted as $\tilde{\boldsymbol{x}}_{2 k}$ and $\tilde{\boldsymbol{x}}_{2 k-1}$, with one arbitrarily designated as the anchor and the other as the positive. Their indices are denoted as $i \in I \equiv\{1 \ldots 2 N\}$ and $j(i)$ respectively. The remaining $2N-2$ samples in $\left\{\tilde{\boldsymbol{x}}_{i}, \tilde{\boldsymbol{y}}_{i}\right\}_{i=1 \ldots 2 N}$ are considered negatives. The self-supervised contrastive loss is defined as follows:
\begin{equation}
\label{ssl}
\mathcal{L}^{\text {self }} = \sum_{i \in I} \mathcal{L}_{i}^{\text {self }} = -\sum_{i \in I} \log \frac{\exp \left(\boldsymbol{z}_{i} \cdot \boldsymbol{z}_{j(i)} / \tau\right)}{\sum_{a \in A(i)} \exp \left(\boldsymbol{z}_{i} \cdot \boldsymbol{z}_{a} / \tau\right)}
\end{equation}
Here, $\boldsymbol{z}_{l}=\varphi_\theta(\tilde{\boldsymbol{x}}_{l})$ denotes the embedding of sample $\tilde{\boldsymbol{x}}_{l}$ outputted by a contrastive network, and $\cdot$ signifies the inner dot product. $A(i) \equiv I \backslash\{i\}$, and $\tau \in \mathcal{R}^{+}$ represents a scalar temperature hyper-parameter.

Nevertheless, the limitation of the loss function described in \autoref{ssl} lies in its inability to effectively utilize label information, rendering it unsuitable for fully-supervised scenarios. To address this issue, a supervised contrastive loss is proposed, building upon the foundation of the self-supervised contrastive loss \cite{Khosla_Teterwak_Wang_Sarna_Tian_Isola_Maschinot_Liu_Krishnan_2020}. In this formulation, samples sharing the same label as each anchor are considered positive pairs. The supervised contrastive loss is defined as follows:
\begin{align}
\label{scl}
\mathcal{L}_{scl}(\varphi) & = \sum_{i \in I} \frac{-1}{|P(i)|} \sum_{p \in P(i)} \log \frac{\exp \left(\varphi(\tilde{\boldsymbol{x}}_{i}) \cdot \varphi(\tilde{\boldsymbol{x}}_{p}) / \tau\right)}{\sum_{a \in A(i)} \exp \left(\varphi(\tilde{\boldsymbol{x}}_{i}) \cdot \varphi(\tilde{\boldsymbol{x}}_{a}) / \tau\right)}
\end{align}

where $P(i) \equiv\left\{p \in A(i): \tilde{\boldsymbol{y}}_{p}=\tilde{\boldsymbol{y}}_{i}\right\}$ denotes the set of indices of all positives of the anchor different from $i$, and $|P(i)|$ represents its cardinality.

\subsection{Balanced Random Forest (BRF) Classifier}
\label{sec:BRF}
The BRF classifier~\cite{Chen2004UsingRF} aims to mitigate the bias towards the majority class that is present in standard random forest algorithms by ensuring that each tree is exposed to a balanced representation of both classes. It can be described as follows:

\begin{enumerate}
  \item \textbf{Sampling}: For each iteration $i$ of the algorithm, where $i = 1, 2, \ldots, N$ and $N$ is the total number of trees to be created in the forest. First, draw a bootstrap sample $S_{\text{minority}}^i$ from the minority class. Then randomly draw, with replacement, the same number of cases from the majority class to form sample $S_{\text{majority}}^i$.
  \item \textbf{Tree Induction}: For each balanced sample $S^i = S_{\text{minority}}^i \cup S_{\text{majority}}^i$, first grow a classification tree $T_i$ to maximum size without pruning, using the CART algorithm~\cite{breiman2017classification}. Then at each node within $T_i$, instead of examining all possible variables $V$ for the optimal split, only a subset of variables $V_{\text{mtry}} \subseteq V$, where $|V_{\text{mtry}}| = \text{mtry}$, is chosen at random, and the search for the optimal split is performed only on $V_{\text{mtry}}$.

  \item \textbf{Aggregation}: After constructing $N$ trees, aggregate their predictions to make the final ensemble prediction $F$. For a given input $x$, the ensemble prediction $F(x)$ can be defined as
$F(x) = \text{mode}\left( \{T_1(x), T_2(x), \ldots, T_N(x)\} \right)$.
  For classification tasks, $\text{mode}$ is the function that selects the most frequently predicted class by the trees.
\end{enumerate}

\section{Methodology}
\label{method}

\begin{algorithm}[t]
    
    \caption{SCLIFD Updates Feature Extractor}\label{alg:Feature Extractor}
    \SetKwFunction{isOddNumber}{isOddNumber}

    \SetKwInOut{KwIn}{Input}
    \SetKwInOut{KwOut}{Output}
    \SetKwInOut{KwReq}{Require}
    \KwIn{Training samples in per-class sets $X^s, \ldots, X^t$}
    \KwReq{Current memory buffer of exemplar sets $\mathcal{P}=\left(P_1, \ldots, P_{s-1}\right)$; Current feature extractor $\varphi_\theta$}
    $\mathcal{D} \leftarrow \bigcup_{y=s, \ldots, t}\left\{(x, y): x \in X^y\right\} \cup \bigcup_{y=1, \ldots, s-1}\left\{(x, y): x \in P_y\right\}$ // Combine training set
    
    ${\varphi_\theta}^{T} \leftarrow \varphi_\theta$  // Freeze pre-update parameters

    \For{$i=1\ldots max\_epoch$}{
    $\mathcal{L}_{encoder}=\sum_{\left(x_i, y_i\right) \in \mathcal{D}}\left[\sum_{y=1}^t \mathcal{L}_{scl}(\varphi_\theta) \right. \left.+\sum_{y=1}^{s-1} \mathcal{L}_{dis}({\varphi_\theta}^T,\varphi_\theta) \right]$

    $\theta_\varphi \leftarrow \theta_\varphi - lr \cdot \frac{\partial \mathcal{L}_{encoder}}{\partial \theta_\mathcal{\varphi}}$ // Update parameters
    } 
    
    \KwOut{Updated feature extractor $\varphi_\theta$}
\end{algorithm}

\begin{algorithm}[t]
    
    \caption{SCLIFD Makes Fault Classification}\label{alg:Classification}
    \SetKwFunction{isOddNumber}{isOddNumber}

    \SetKwInOut{KwIn}{Input}
    \SetKwInOut{KwOut}{Output}
    \SetKwInOut{KwReq}{Require}
    \KwIn{Samples to be classified $X$}
    \KwReq{Current memory buffer of exemplar sets $\mathcal{P}=\left(P_1, \ldots, P_{t}\right)$; Current feature extractor $\varphi_{\theta}$; BRF classifier $\mathcal{C}$}

    $\mathcal{D} \leftarrow \bigcup_{y=1, \ldots, t}\left\{(x, y): x \in P_y\right\}$ 
    // Combine previous exemplar sets
    
    $ \{z, y\} \in \mathcal{Z} \leftarrow \varphi_{\theta}(\mathcal{D})$ // Extract features from previous exemplars

    $\mathcal{C}^{'} \leftarrow$Train $\mathcal{C}$ using $\mathcal{Z}$ as \autoref{sec:BRF} described

    $Y^{*} \leftarrow \mathcal{C}^{'}(\varphi_{\theta}(X))$ // Make classifications using extracted features

    \KwOut{Predicted classifications  $Y^{*}$}
\end{algorithm}

In the upcoming subsections, we will present the principal process of class incremental fault diagnosis through SCLIFD and provide a detailed explanation of each stage.
\subsection{Overview of Class Incremental Fault Diagnosis}
\label{overview}
SCLIFD class-incrementally diagnoses limited fault data consisting of four steps, as shown in \autoref{alg:train}:
\begin{enumerate}
    \item \textbf{Process Data.} 
In each session, SCLIFD is exposed to samples from new classes. This new class data is then merged with an exemplar set, which is created using the MES method (refer to \autoref{alg:Feature Extractor} and \autoref{sec:herding}). After combining, the data undergoes augmentation and serves to further train the feature extraction network, allowing it to adjust to the new class data.
    \item \textbf{Train the Feature Extraction Network.} We use supervised contrastive knowledge distillation to update the feature extractor in each incremental session, as illustrated in  \autoref{sec:SCLKD}, \autoref{update} and \autoref{alg:Feature Extractor}. It consists of \autoref{scl} $\mathcal{L}_{scl}$ and \autoref{dis} $\mathcal{L} _ { dis }$. The reason is that SCL can largely enhance SCLIFD's feature extraction capability. 
    Furthermore, knowledge distillation can alleviate catastrophic forgetting because the feature extraction capability of old classes can be distilled from the feature extractor in the last session to the extractor in the current session.
    \item \textbf{Construct the Exemplar Set.} This stage constructs a $K$ size memory buffer. It includes exemplar sets of a small number of previous classes' samples selected by the MES strategy (see \autoref{sec:herding}). The memory buffer will be used in the initial data process stage with the new class data in the next incremental session. This is called sample replay, which can alleviate catastrophic forgetting and is widely used in class incremental learning \cite{icarl, Wu_2019_CVPR, Castro_2018_ECCV}.
    \item \textbf{Train Fault Classifer.}  We utilize the BRF classifier~\cite{Chen2004UsingRF} to alleviate the imbalance between normal class and fault classes, as described in \autoref{sec:BRF}. As shown in \autoref{alg:Classification},
    the feature extraction network initially extracts features from the exemplar memory buffer, which includes exemplars from both old and novel classes. Then the extracted features and their labels are used to train the classifier. 
    \item \textbf{Online Fault diagnosis.} First, the updated feature extraction network will be used to extract the feature representations of the online samples. Then the trained classifier makes online fault diagnosis classifications on features extracted from the input samples using the current feature extractor.
    \item \textbf{Periodic Offline Training for Model Updating.} In the real-world industry, fully online training for class incremental fault diagnosis is challenging, especially for new fault categories. To address this, we propose a periodic offline training strategy, where the model is periodically retrained offline with accumulated fault data, including newly emerged fault classes. After retraining, the updated model is deployed online to diagnose new faults in real time. This approach keeps a balance between system performance and real-world deployment constraints.
\end{enumerate}

\begin{table*}[t]
\scriptsize
\centering
\caption{TEP and MFF Dataset Setting in Training and Testing Process}
\label{datasets}
\begin{tabular}{@{}ccccccclccc@{}}
\toprule
\multirow{2}{*}{Dateset} & \multirow{2}{*}{Mode} & \multirow{2}{*}{\makecell{Total\\ Classes}} & \multirow{2}{*}{\makecell{Incremental \\ Sessions}} & \multirow{2}{*}{\makecell{Novel-Class\\ Shot}} & \multicolumn{2}{c}{Training  Set} &  & \multicolumn{2}{c}{Testing Set} & \multirow{2}{*}{Memory Buffer $K$} \\ \cmidrule(lr){6-7} \cmidrule(lr){9-10}
 &  &  &  &  & Normal Class & Fault Class &  & Normal Class & Fault Class &  \\ \midrule
\multirow{2}{*}{TEP} & Imbalanced & \multirow{2}{*}{10} & \multirow{2}{*}{5} & \multirow{2}{*}{2} & \multirow{2}{*}{500} & 48 &  & \multirow{2}{*}{800} & \multirow{2}{*}{800} & 100 \\
 & Long-Tailed &  &  &  &  & 20 &  &  &  & 40 \\ \midrule
\multirow{2}{*}{MFF} & Long-Tailed 1 & \multirow{2}{*}{5} & \multirow{2}{*}{5} & \multirow{2}{*}{1} & \multirow{2}{*}{200} & 10 &  & \multirow{2}{*}{800} & \multirow{2}{*}{800} & 10 \\
 & Long-Tailed 2 &  &  &  &  & 5 &  &  &  & 5 \\ \bottomrule
\end{tabular}
  \begin{tablenotes}    
        \footnotesize               
        \item[1]  The number of fault class samples ranges from 20 to 50
        for imbalanced fault diagnosis and ranges from 1 to 20 for long-tailed fault diagnosis. \cite{Peng_Lu_Tao_Ma_Zhang_Wang_Zhang_2022}
    \end{tablenotes}            
\end{table*}

\subsection{Update Feature Extractor by Supervised Contrastive Knowledge Distillation}
\label{sec:SCLKD}
As illustrated in \autoref{update} and \autoref{alg:Feature Extractor}, we propose supervised contrastive knowledge distillation to update the feature extractor in each incremental session.
We have introduced SCL in detail in \autoref{sec:SCLKD}.
$\mathcal{L}_{scl}$ loss
is obtained from the output features of the current feature extractor to enhance SCLIFD's feature extraction ability. 
We will explain how we integrate SCL with Knowledge Distillation (KD) to train the feature extractor. It can alleviate catastrophic forgetting because it can effectively enhance the performance of the feature extractor in the current session for old classes under the supervision of the one in the last session. 

Specifically, let 
$a \in A(i) \equiv I \backslash\{i\}$. 
Then each $\tilde{\boldsymbol{x}}_{i}$ and $\boldsymbol{x}_{a}$ are mapped by the feature extractor network $\varphi_\theta$ and normalized into feature vector representations $\boldsymbol{z}_{i}$ and $\boldsymbol{z}_{a}$, i.e., $\boldsymbol{z}=\varphi_\theta(\boldsymbol{x})$. 
$P(\boldsymbol{z}_{i}; \boldsymbol{z}_{a})$ denotes the similarity score between $\boldsymbol{z}_{i}$ and $\boldsymbol{z}_{a}$. Then the softmax function is applied with the temperature scaling factor $\tau$ to $P(\boldsymbol{z}_{i}; \boldsymbol{z}_{a})$, resulting in:
\begin{align}
P(\boldsymbol{z}_{i}; \boldsymbol{z}_{a})  & = \frac{\exp \left(\boldsymbol{z}_{i} \cdot \boldsymbol{z}_{a} / \tau\right)}{\sum_{j \in A(i)} \exp \left(\boldsymbol{z}_{i} \cdot \boldsymbol{z}_{j} / \tau\right)}
\end{align}
Next, the last session's feature network is frozen to be utilized as the teacher network. We adopt cross-entropy loss to obtain the KD loss, which is defined as follows:

{\scriptsize
\begin{equation}
\label{dis}
\begin{split}
    \mathcal{L} _ { dis }({\varphi}^{T},\varphi) &=\frac{1}{2 N} \sum_{i \in I} \sum_{a \in A(i)} \mathcal{L}_{i, a}^{d i s} \\
&=-\frac{1}{2 N} \sum_{i \in I} \sum_{a \in A(i)} P(\boldsymbol{z}_{i}; \boldsymbol{z}_{a})^{T} \log \left(P(\boldsymbol{z}_{i}; \boldsymbol{z}_{a})^{S}\right) \\
&=-\frac{1}{2 N} \sum_{i \in I} \sum_{a \in A(i)} P(\varphi^{T}(\tilde{\boldsymbol{x}}_{i}); \varphi^{T}(\boldsymbol{x}_{a})) \log \left(P(\varphi(\tilde{\boldsymbol{x}}_{i}); \varphi(\boldsymbol{x}_{a})\right) 
\end{split}
\end{equation}}

In this way, we make feature space distillation for the current feature extractor, which emulates the latent feature space of the teacher. This is distinctive from the previous work~\cite{icarl} which makes response distillation~\cite{Gou_Yu_Maybank_Tao_2021}, i.e., classification logits distillation from a teacher network.

As demonstrated in \autoref{alg:Feature Extractor}, $\mathcal{L}_{scl}$ is merged with $\mathcal{L}_{dis}$ to form the ultimate $\mathcal{L}_{encoder}$. This is utilized to update the feature network during each incremental session. The revised feature network is then employed in the ultimate stage of online fault diagnosis.

Our Supervised Contrastive Knowledge Distillation extends standard SCL by introducing cross-session knowledge transfer. Instead of only learning better features within a session, we distill knowledge from the previous session's feature extractor to the current one, aiding in retaining past knowledge and reducing catastrophic forgetting. Unlike traditional distillation, which focuses on class logits, our method distills feature representations, aligning the new feature extractor with the old one. This component ensures that features for both old and new classes remain preserved and discriminative.

\subsection{Prioritized Exemplar Selection Method MES}
\label{sec:herding}
Our method adopts a sample replay strategy to overcome the problem of catastrophic forgetting. Namely, at each incremental session, we select a subset of each incremental class to put it into the memory buffer, which is used to update the feature extractor with both old and new class samples. As a result, we propose the Marginal Exemplar Selection (MES) strategy to select an exemplar subset consisting of diverse and marginal samples. The goal of the MES strategy is to identify and select samples that are positioned on the edges within the feature space, essentially those that are likely to lie on the boundaries of the data distribution. By choosing such samples, the model is encouraged to develop a deeper understanding of the boundaries within the data distribution. We show the procedures of MES strategy in \autoref{alg:train}.

Additionally, as new classes of data are continuously added and stored for future training, both computational and storage demands grow. To address this issue, our approach maintains a constant number of $K$ exemplars in a memory buffer for storing both old and new class data, aligning with techniques from prior research~\cite{icarl, Wu_2019_CVPR, Castro_2018_ECCV}. Given that $t$ classes have been encountered so far, our method allocates storage of $m=K/t$ exemplars per class.
Assuming the new class comprises a set $X = \{x_1, \ldots, x_n\}$ of class $y$, and the current feature extractor is $\varphi_\theta: \mathcal{X} \rightarrow \mathbb{R}^d$, the class mean $\mu$ is calculated as follows:
\begin{equation}
 \label{eq:mean}
\mu = \frac{1}{n} \sum_{x \in X} \varphi_\theta(x)
 \end{equation}
MES iteratively selects and stores exemplars $p_{1}, \ldots, p_{m}$ until the number of exemplars in the set matches the target number $m$. The selection process is computed in the following way:
 \begin{equation}
 \label{argmax}
     p_{k} = \underset{x \in X}{\operatorname{argmax}}\parallel \mu-\frac{1}{k}[\varphi_\theta(x)+\sum_{j = 1}^{k-1} \varphi_\theta\left(p_{j}\right)]\parallel 
 \end{equation}
where $k$ is iterated from 1 to $m$. This process results in the exemplar set becoming a prioritized set.

The MES strategy is crucial for addressing catastrophic forgetting. Unlike Herding~\cite{welling2009herding}, which selects samples near class centroids, MES prioritizes marginal samples close to decision boundaries, improving class distinction and generalization. This focus ensures minority classes are well-represented, balancing class representation and preventing bias toward majority classes. As a result, MES enhances boundary recognition and boosts incremental learning performance. In \autoref{sec:ablation}, we will also prove that the model adopting the MES strategy performs better than the one adopting mixed samples from both the MES and Herding strategies.

\section{Experiment Study}

\label{Experiment}

We conduct comprehensive experiments on two datasets: the benchmark TEP dataset \cite{LAWRENCERICKER1996205} and the practical dataset MFF \cite{ruiz2015statistical}, to assess the efficacy of our SCLIFD. Due to the limited open-sourced models in the fault diagnosis domain, the performance of SCLIFD is compared to some classical and state-of-the-art methods in computer vision domains. 
Following previous works~\cite{Li2018,song2023learning,icarl,Hell_Pestana_Soares_Goliatt_2022}, we choose the classification accuracy and average accuracy in all incremental sessions as our evaluation metrics.
Moreover, ablation experiments are also performed to prove the effectiveness of each component.

\begin{table}[t]
\centering
\caption{Descriptions of the Faults in the TEP Dataset~\cite{LAWRENCERICKER1996205}}
\label{tab1}
\begin{tabular}{cc}
\toprule
Fault & Description    \\
\midrule
1 &    Step in A/C feed ratio (Stream 4)   \\

2 & Step in B composition(Stream 4)   \\

3 &    Step in D Feed temperature(Stream 3)  \\

4 &    Step in reactor cooling water inlet temperature  \\

5 &    Step in condenser cooling water inlet temperature   \\

6 &    Step in A Feed loss in Stream 1   \\

7 &    Stream 4 Header pressure loss  \\

8 &    Random variations in A,B,C compositions(Stream 4)  \\

9 &    Random variations in D feed temperature  \\

10 & Random variations in C feed temperature   \\

11 & Random variations in reactor cooling water inlet temperature   \\

12 & Random variations in condenser cooling water inlet temperature  \\

13 & Slow drift in reaction kinetics  \\

14 & Reactor cooling water valve sticking   \\

15 & Condenser cooling water valve sticking  \\

16-20 & Unknown    \\

21 & Valve for stream 4 fixed at the steady-state position\\
\bottomrule
\end{tabular}
\end{table}

\subsection{Implementation Details}
\subsubsection{Datasets}
TEP dataset~\cite{LAWRENCERICKER1996205} is widely recognized in the fault diagnosis field for its simulation of realistic chemical processes. It comprises 52 variables, including temperature and pressure continually monitored by sensors. It features 20 types of faults, as shown in \autoref{tab1}. We have selected nine fault types (type 1, 2, 4, 6, 7, 8, 12, 14, and 18) along with one normal type (type 0) at random, to demonstrate the effectiveness of SCLIFD.
Access to the dataset is available through \href{https://depts.washington.edu/control/LARRY/TE/download.html}{the provided link}.
The MFF~\cite{ruiz2015statistical} dataset originates from the Three-phase Flow Facility system at Cranfield University and consists of 24 process variables sampled at 1 Hz. It has 6 types of faults consisting of 23 variables collected from a real-world multiphase flow facility. 
We randomly pick faults 1, 2, 3, and 4 and normal class 0 in the experiments. It can be downloaded from \href{https://www.mathworks.com/matlabcentral/fileexchange/50938-a-benchmark-case-for-statistical-process-monitoring-cranfield-multiphase-flow-facility}{the link}.

\begin{table}[t]
\scriptsize
\setlength\tabcolsep{2pt}
\centering
\caption{Experiment Results Comparison on TEP and MFF Dataset}
\label{tab:baseline}
\begin{tabular}{@{}ccccccccc@{}}
\toprule
\multirow{2}{*}{Dataset} & \multirow{2}{*}{Diagnosis Mode} & \multirow{2}{*}{Method} & \multicolumn{6}{c}{Accuracy(\%) in All Incremental  Sessions$\uparrow$} \\ \cmidrule(l){4-9} 
 &  &  & 1 & 2 & 3 & 4 & 5 & Average \\ \midrule
\multirow{18}{*}{TEP} & \multirow{9}{*}{Imbalanced} & LwF.MC & \textbf{99.43} & 42.92 & 18.89 & 13.96 & 10.76 & 37.19 \\
 &  & Finetuning & 99.38 & 47.02 & 32.00 & 15.87 & 17.63 & 42.38 \\
 &  & iCaRL & 98.69 & 72.41 & 60.67 & 58.26 & 54.02 & 68.81 \\
 &  & EEIL & 58.12 & 33.24 & 17.67 & 14.87 & 11.24 & 27.03 \\
 &  & BiC & 59.55 & 32.57 & 25.35 & 19.63 & 17.76 & 30.97 \\
 &  & SAVC & 89.22 & 59.87 & 50.54 & 36.50 & 30.85 & 53.40 \\
 &  & WaRP-CIFSL & 98.35 & 55.54 & 52.48 & 42.59 & 37.84 & 57.36 \\
 &  & BiDistFSCIL & 99.20 & 71.93 & 63.35 & 50.96 & 47.98 & 66.68 \\
 &  & \textbf{Ours} & 98.89  & \textbf{98.48}           & \textbf{98.29}  & \textbf{83.16}  & \textbf{72.25} & \textbf{90.23} \\ \cmidrule(l){2-9} 
 & \multirow{9}{*}{Long-Tailed} & LwF.MC & \textbf{99.32} & 34.43 & 26.49 & 14.57 & 13.27 & 37.62 \\
 &  & Finetuning & 99.09 & 46.79 & 31.75 & 14.19 & 15.63 & 41.49 \\
 &  & iCaRL & 98.35 & 70.21 & 58.17 & 54.12 & 51.63 & 66.49 \\
 &  & EEIL & 57.81 & 27.14 & 13.38 & 10.26 & 8.14 & 23.35 \\
 &  & BiC & 52.66 & 27.93 & 18.67 & 13.65 & 10.46 & 24.67 \\
 &  & SAVC & 97.24 & 56.35 & 42.11 & 32.21 & 25.84 & 50.75 \\
 &  & WaRP-CIFSL & 97.27 & 51.31 & 44.40 & 39.01 & 36.68 & 53.73 \\
 &  & BiDistFSCIL & 99.09 & 68.69 & 54.94 & 46.01 & 43.01 & 62.35 \\
 &  & \textbf{Ours} & 98.86 & \textbf{93.21} & \textbf{94.05} & \textbf{72.33} & \textbf{62.89} & \textbf{84.27} \\ \midrule
\multirow{18}{*}{MFF} & \multirow{9}{*}{Long-Tailed 1} & LwF.MC & 100.00 & 80.68 & 43.91 & 38.30 & 28.85 & 58.35 \\
 &  & Finetuning & 100.00 & 50.00 & 33.33 & 25.00 & 20.00 & 45.67 \\
 &  & iCaRL & 100.00 & 94.19 & 92.50 & 97.66 & 93.20 & 95.51 \\
 &  & EEIL & 100.00 & 43.34 & 33.87 & 25.93 & 22.19 & 45.07 \\
 &  & BiC & 100.00 & 48.45 & 34.25 & 27.81 & 19.23 & 45.95 \\
 &  & SAVC & 100.00 & 99.94 & 70.50 & 51.78 & 38.80 & 72.20 \\
 &  & WaRP-CIFSL & 100.00 & 97.69 & 98.42 & 96.44 & 96.78 & 97.86 \\
 &  & BiDistFSCIL & 100.00 & 98.19 & 98.63 & 96.72 & 96.18 & 97.94 \\
 &  & \textbf{Ours} &  \textbf{100.00} & \textbf{100.00} & \textbf{100.00}           & \textbf{100.00}  & \textbf{98.65} & \textbf{99.73} \\ \cmidrule(l){2-9} 
 & \multirow{9}{*}{Long-Tailed 2} & LwF.MC & 100.00 & 58.47 & 40.20 & 52.38 & 42.31 & 58.67 \\
 &  & Finetuning & 100.00 & 50.00 & 33.33 & 25.00 & 20.00 & 45.67 \\
 &  & iCaRL & 100.00 & 74.94 & 82.13 & 84.56 & 89.93 & 86.31 \\
 &  & EEIL & 100.00 & 50.12 & 31.62 & 24.56 & 21.25 & 45.51 \\
 &  & BiC & 100.00 & 41.74 & 30.10 & 21.53 & 17.12 & 42.10 \\
 &  & SAVC & 100.00 & 93.56 & 95.67 & 72.13 & 56.13 & 83.50 \\
 &  & WaRP-CIFSL & 100.00 & 100.00 & 98.88 & 98.88 & 96.60 & 98.87 \\
 &  & BiDistFSCIL & 100.00 & 99.62 & 100.00 & 96.91 & 88.90 & 97.09 \\
 &  & \textbf{Ours} & \textbf{100.00} & \textbf{100.00} & \textbf{100.00}           & \textbf{98.03} & \textbf{99.15} & \textbf{99.44} \\ \bottomrule
\end{tabular}

\end{table}
Moreover, \autoref{datasets} shows the number of novel classes added at each incremental session for TEP and MFF datasets, the total incremental sessions, training and testing set size for normal and fault classes, and memory buffer size $K$. 
To simulate limited fault data, we restrict the number of samples for fault classes. Following \cite{Peng_Lu_Tao_Ma_Zhang_Wang_Zhang_2022}, fault class samples range from 20 to 50 for imbalanced diagnosis and from 1 to 20 for long-tailed diagnosis. In the TEP dataset, fault classes have 48 or 20 samples, while in the MFF dataset, they range from 10 to 5. Class imbalance is created by having far more samples in the normal class than fault classes. The long-tail distribution is simulated with fault classes having very few samples, especially in the MFF dataset, with two different distributions resulting in samples ranging from 10 to 5.

\subsubsection{Compared Methods}
We compare our method with several classical and state-of-the-art approaches, including LwF.MC \cite{Li2018}, Finetuning \cite{icarl}, iCaRL \cite{icarl}, EEIL \cite{Castro_2018_ECCV}, BiC \cite{Wu_2019_CVPR},
which serve as widely classical methods in class incremental learning, and
SAVC \cite{song2023learning}, WaRP-CIFSL \cite{kim2023warping}, and BiDistFSCIL \cite{zhao2023few}, which represent the latest advances. LwF.MC \cite{Li2018} preserves performance on old tasks without accessing their data. iCaRL \cite{icarl} combines a mean-of-exemplars classifier, sample selection, and knowledge distillation to retain old knowledge. Finetuning \cite{icarl} continues training but lacks mechanisms to prevent forgetting. EEIL \cite{Castro_2018_ECCV} employs cross-distillation loss and memory units for end-to-end incremental learning. BiC \cite{Wu_2019_CVPR} addresses classification bias via a linear bias correction layer. SAVC \cite{song2023learning} uses virtual classes to enhance class separation. WaRP-CIFSL \cite{kim2023warping} rotates the weight space to consolidate old knowledge, and BiDistFSCIL \cite{zhao2023few} combines dual teacher models to mitigate overfitting and forgetting.

\subsubsection{Parameter Setting}
In the representation learning stage, we use ResNet18 as the feature extractor. We train the network using Adam optimizer with batches of 512 samples, a weight decay of $10^{-5}$, a temperature $\tau$ of 0.07, and a learning rate of 0.01 in 500 epochs. For data augmentation in SCL, we randomly select a segment of the input sequence, shuffle this segment's order, and integrate it back into the original sequence. BRF classifier is implemented using the \href{https://imbalanced-learn.org/stable/references/generated/imblearn.ensemble.BalancedRandomForestClassifier.html#balancedrandomforestclassifier}{imbalanced-learn} library.

\begin{figure*}[htbp]
    \centering
    \includegraphics[width=\textwidth]{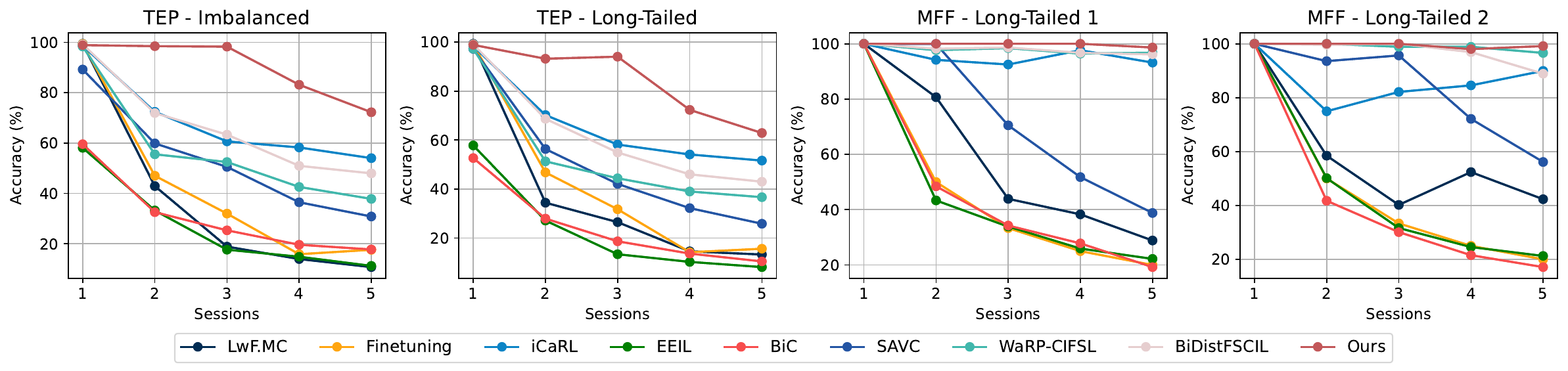}
    \caption{Experiment result comparisons of different methods on TEP and MFF Dataset.}
    \label{fig:baseline}
\end{figure*}

\begin{figure*}[t]
    \centering
    \includegraphics[width=\textwidth]{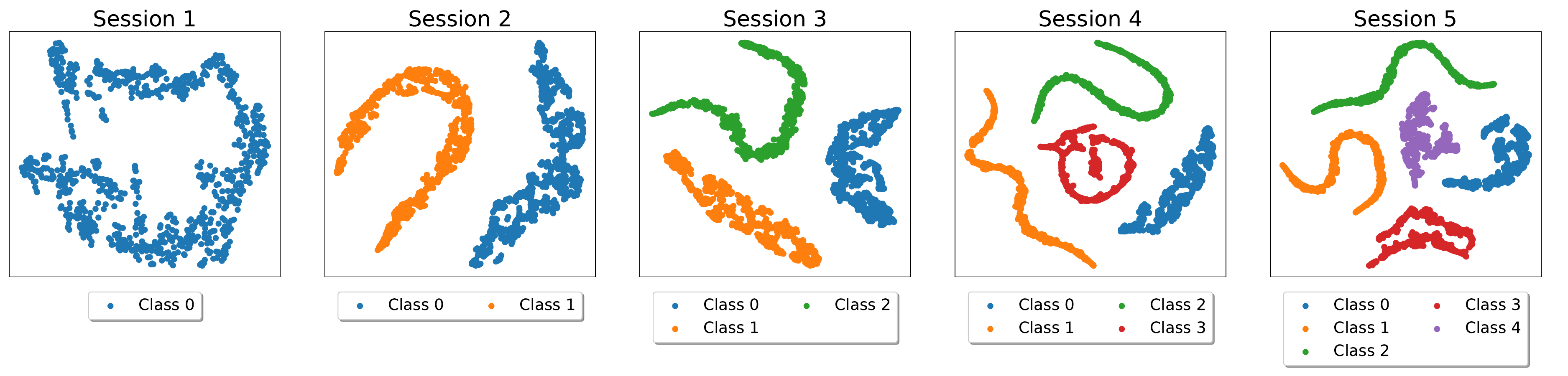}
    \caption{
    T-SNE visualization of each class's features extracted by SCLIFD. It extracts discriminative features for different classes over multiple sessions.}
    \label{fig:tsne_subplots}
\end{figure*}

\begin{table*}[t]
\scriptsize
\centering
\caption{Ablation Results on TEP and MFF Dataset}
\label{tab:ablation}
\begin{tabular}{@{}ccccclcccccc@{}}
\toprule
\multirow{2}{*}{Dataset} & \multirow{2}{*}{Diagnosis Mode} & \multicolumn{3}{c}{Components} &  & \multicolumn{6}{c}{Accuracy in All Incremental Sessions$\uparrow$ (Bold for the Best One)} \\ \cmidrule(lr){3-5} \cmidrule(l){7-12} 
 &  & SCL & MES & BRF &  & 1 & 2 & 3 & 4 & 5 & Average \\ \midrule
 \multirow{20}{*}{TE} &  \multirow{10}{*}{Imbalanced} & & & & & 65.17\% & 58.42\% & 51.65\% & 40.76\% & 34.78\% & 50.16\% \\
 & & $\checkmark$ &  &  &  & 79.77\% & 93.57\% & 92.70\% & 72.21\% & 65.72\% & 80.80\% \\
 & &  & $\checkmark$ &  &  & 69.03\% & 72.92\% & 78.79\% & 67.21\% & 64.19\% & 70.43\% \\
 & &  &  & $\checkmark$ &  & 98.52\% & 69.35\% & 53.04\% & 50.81\% & 39.77\% & 62.30\% \\
 & & $\checkmark$ & $\checkmark$ & & & 77.10\% & 94.91\% & 91.13\% & 76.66\% & 67.16\% & 81.39\% \\
 & & $\checkmark$ & Herding & $\checkmark$ & & 99.09\% & 97.29\% & 96.63\% & 80.99\% & 67.73\% & 88.35\% \\
 & & $\checkmark$ & Mixed & $\checkmark$ &  & 98.92\% & 97.86\% & 97.08\% & 83.40\% & 70.88\% & 89.63\% \\
 & & $\checkmark$ & & $\checkmark$ & & 99.09\% & 82.95\% & 67.52\% & 48.67\% & 43.95\% & 68.44\% \\
 & & & $\checkmark$ & $\checkmark$ & & 98.92\% & 73.60\% & 74.62\% & 63.23\% & 59.87\% & 74.05\% \\
 & & $\checkmark$ & $\checkmark$ & $\checkmark$ & & 98.98\% & 98.48\% & 98.29\% & 83.16\% & 72.25\% & 90.23\% \\ \cmidrule(l){2-12}  
 & \multirow{10}{*}{Long-Tailed} & & & & & 59.26\% & 46.19\% & 34.96\% & 29.62\% & 26.94\% & 39.39\% \\
 & & $\checkmark$ &  &  &  & 64.49\% & 89.08\% & 86.59\% & 63.78\% & 55.75\% & 71.94\% \\
 & &  & $\checkmark$ &  &  & 63.13\% & 64.46\% & 65.99\% & 52.88\% & 51.00\% & 59.49\% \\
 & &  &  & $\checkmark$ &  & 81.70\% & 61.25\% & 61.11\% & 46.43\% & 32.45\% & 56.59\% \\
 & & $\checkmark$ & $\checkmark$ & & & 66.31\% & 90.42\% & 86.57\% & 62.84\% & 61.00\% & 73.43\% \\
 & & $\checkmark$ & Herding & $\checkmark$ & & 98.86\% & 89.58\% & 89.03\% & 68.20\% & 60.48\% & 81.23\% \\
 & & $\checkmark$ & Mixed & $\checkmark$ &  & 99.09\% & 85.45\% & 90.71\% & 67.15\% & 57.01\% & 79.88\% \\
 & & $\checkmark$ & & $\checkmark$ & & 98.86\% & 68.90\% & 53.91\% & 37.42\% & 32.27\% & 58.27\% \\
 & & & $\checkmark$ & $\checkmark$ & & 98.41\% & 69.73\% & 57.34\% & 51.28\% & 48.31\% & 65.01\% \\
 & & $\checkmark$ & $\checkmark$ & $\checkmark$ & & 98.86\% & 93.21\% & 94.05\% & 72.33\% & 62.89\% & 84.27\% \\ \midrule
 \multirow{20}{*}{MFF} & \multirow{10}{*}{Long-Tailed 1} & & & & & 100.00\% & 60.94\% & 59.83\% & 43.06\% & 33.40\% & 59.45\% \\
& & $\checkmark$ &  &  &  & 100.00\% & 87.81\% & 99.33\% & 81.63\% & 81.10\% & 89.97\% \\
& &  & $\checkmark$ &  &  & 100.00\% & 99.69\% & 98.88\% & 81.25\% & 81.48\% & 92.26\% \\
& &  &  & $\checkmark$ &  & 100.00\% & 90.56\% & 94.58\% & 89.63\% & 78.55\% & 90.66\% \\
 & & $\checkmark$ & $\checkmark$ & & & 100.00\% & 100.00\% & 100.00\% & 97.46\% & 88.07\% & 97.11\% \\
 & & $\checkmark$ & Herding & $\checkmark$ & & 100.00\% & 93.38\% & 90.75\% & 98.31\% & 92.78\% & 95.04\% \\
 & & $\checkmark$ & Mixed & $\checkmark$ &  & 100.00\% & 100.00\% & 100.00\% & 100.00\% & 89.70\% & 97.94\% \\
 & & $\checkmark$ & & $\checkmark$ & & 100.00\% & 99.25\% & 91.58\% & 83.09\% & 86.35\% & 92.05\% \\
 & & & $\checkmark$ & $\checkmark$ & & 100.00\% & 100.00\% & 99.67\% & 97.09\% & 92.65\% & 97.88\% \\
 & & $\checkmark$ & $\checkmark$ & $\checkmark$ & & 100.00\% & 100.00\% & 100.00\% & 100.00\% & 98.65\% & 99.73\% \\ \cmidrule(l){2-12} 
 & \multirow{10}{*}{Long-Tailed 2} & & & & & 100.00\% & 55.88\% & 41.13\% & 39.47\% & 16.88\% & 50.67\% \\
& & $\checkmark$ &  &  &  & 100.00\% & 73.31\% & 64.88\% & 68.16\% & 49.90\% & 71.25\% \\
& &  & $\checkmark$ &  &  & 100.00\% & 99.00\% & 99.96\% & 99.88\% & 89.70\% & 97.71\% \\
& &  &  & $\checkmark$ &  & 100.00\% & 69.81\% & 66.00\% & 66.31\% & 67.50\% & 73.93\% \\
 & & $\checkmark$ & $\checkmark$ & & & 100.00\% & 96.88\% & 91.67\% & 94.94\% & 92.50\% & 95.20\% \\
 & & $\checkmark$ & Herding & $\checkmark$ & & 100.00\% & 98.31\% & 98.83\% & 90.16\% & 93.33\% & 96.13\% \\
 & & $\checkmark$ & Mixed & $\checkmark$ &  & 100.00\% & 99.13\% & 100.00\% & 81.66\% & 86.45\% & 93.45\% \\
 & & $\checkmark$ & & $\checkmark$ & & 100.00\% & 69.69\% & 92.04\% & 67.19\% & 48.23\% & 75.43\% \\
 & & & $\checkmark$ & $\checkmark$ & & 100.00\% & 95.50\% & 72.58\% & 70.66\% & 73.80\% & 82.51\% \\
 & & $\checkmark$ & $\checkmark$ & $\checkmark$ & & 100.00\% & 100.00\% & 100.00\% & 98.03\% & 99.15\% & 99.44\% \\ \bottomrule
\end{tabular}
\end{table*}

\subsection{Experimental Results}
\subsubsection{Comparative Results}
This section presents the comparative results of various methods applied to imbalanced and long-tailed fault diagnosis using the TEP and MFF datasets. The trends in accuracy across all incremental sessions for the different methods are depicted in \autoref{fig:baseline} and \autoref{tab:baseline}.
The experimental results are summarized as follows:

\noindent
\textbf{Superior accuracy across all imbalance ratios.}\ For each fault diagnosis case of different imbalance ratios, the average classification accuracies of our SCLIFD always outperform other state-of-the-art methods. For example, through all class incremental sessions of the TEP dataset, SCLIFD achieves the average accuracies of 90.23\% and 84.27\% under the imbalanced and long-tailed fault diagnosis cases respectively. 

\noindent
\textbf{Effectively reduces catastrophic forgetting.}\
 As incremental sessions proceed, the accuracies of SCLIFD are continuously higher than those of other methods. For example, in session 5 of the imbalanced case under the TEP dataset, the accuracy of SCLIFD is 72.25\%, which is 18.23\% higher than the second-best iCaRL of 54.02\% accuracies. Although the accuracies of LwF.MC is the highest in session 1 under the TEP dataset, our method greatly outperforms it in subsequent incremental sessions. 
This demonstrates that our method effectively alleviates catastrophic forgetting.

\subsubsection{Further Analyses}
In this section, we further analyze the reasons behind the improved performance of our method.

\noindent
\textbf{Continuously extracts discriminative features.}\
As shown in \autoref{fig:tsne_subplots}, we plot the T-SNE visualizations of each class’s features in each incremental session under the MFF's imbalanced case. It illustrates the distribution and separation of feature vectors extracted by SCLIFD. It also proves the model effectively and continuously extracts discriminative features as the learning classes increase.

\begin{figure}
    \centering
    \includegraphics[width=\linewidth]{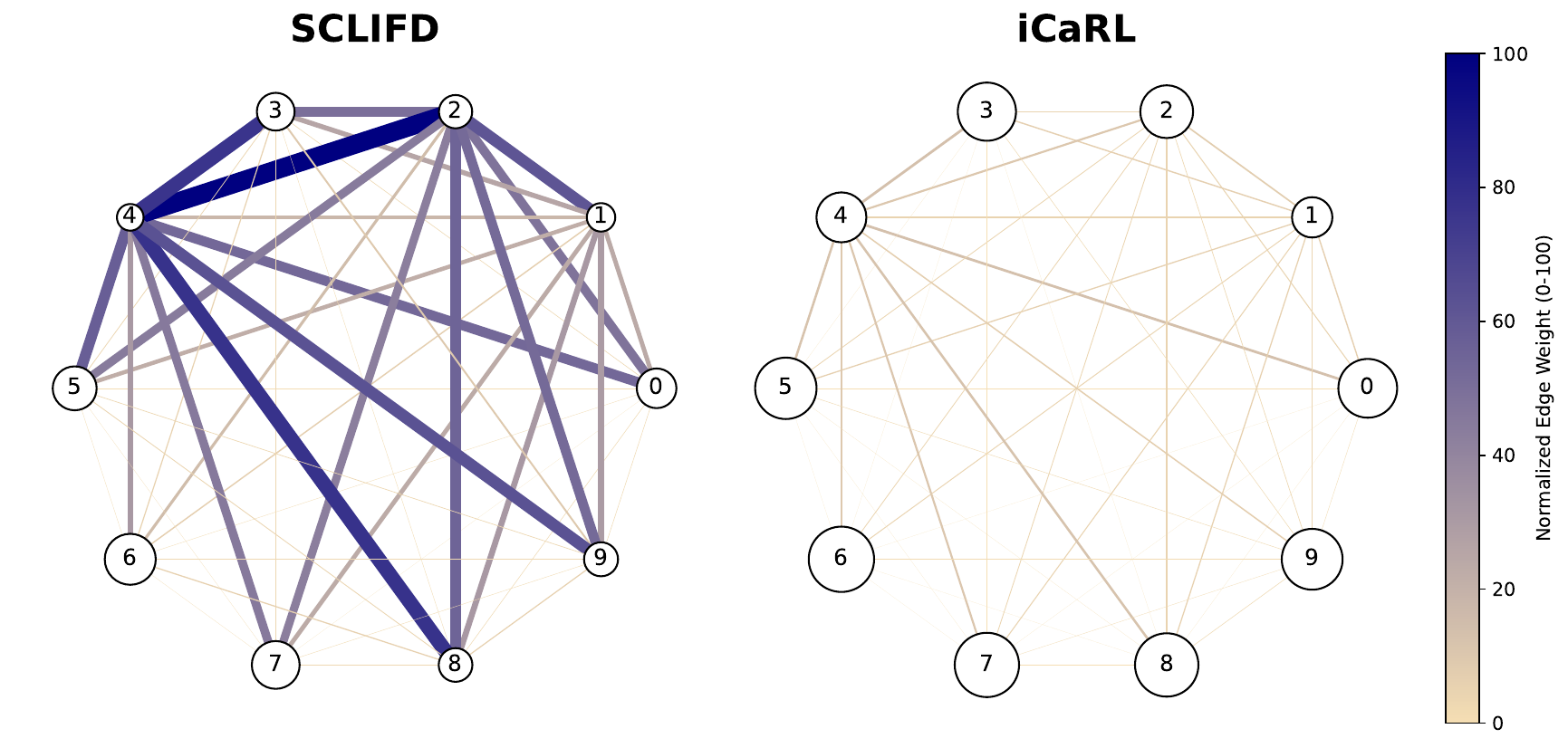}
    \caption{The node network graph for features extracted by our SCLIFD and iCaRL on the imbalanced TEP dataset. \textbf{Smaller nodes and thicker, darker edges indicate clearer class separation and improved feature distinguishability for the classifier.}}
    \label{fig:network}
\end{figure}

\noindent \textbf{Tighter intra-class, larger inter-class separations.}\
To further demonstrate the superiority of our feature extractor, we compute the inter-class (KL divergence between two-class pairs) and intra-class distances (KL divergence within the same class) to create a class network graph on the imbalanced TEP dataset for our method and the second-best method, iCaRL. In \autoref{fig:network}, each node represents a class, with node size indicating intra-class distance and edges representing inter-class distance. Smaller nodes in SCLIFD show tighter clustering within classes compared to iCaRL, while thicker, darker edges indicate greater separation between classes, enabling better class distinction.

\noindent \textbf{Interplay Between Components.}
The components of our framework are designed to work cohesively, ensuring their combined effect surpasses individual contributions:
\begin{itemize}  
\item \textbf{SCKD enhances feature quality}, enabling MES to effectively identify boundary samples. High-quality feature representations help locate the most challenging samples (those near decision boundaries), which MES selects for memory storage, ensuring the buffer contains the most informative samples for incremental updates.
\item \textbf{MES ensures a balanced and meaningful memory buffer}, directly benefiting the BRF classifier. By prioritizing boundary samples, MES trains the BRF on a challenging and representative subset of data, improving decision-making in imbalanced scenarios.
\item \textbf{BRF leverages the improved feature space created by SCKD and MES.} The classifier achieves tighter intra-class clustering and wider inter-class separations, resulting in more accurate classifications, especially for minority classes.
\end{itemize}


\subsection{Ablation Study}
\label{sec:ablation}

Our approach mainly consists of three components, which are SCL, MES, and BRF classifier. We analyze the effect of individual components on TEP and MFF datasets under different imbalance ratios by ablating each part. First, we have conducted experiments to verify the effects of using the SCL, MES, and BRF modules individually.
Additionally, the methods without the SCL module adopt the cross-entropy loss and distillation loss as iCaRL~\cite{icarl} does. Moreover, in terms of the sample replay method, the ablation experiments adopt random selection by default and compare Herding~\cite{welling2009herding}, as well as a mixed approach where half the samples are selected via Herding and the other half using MES, against the MES module. Finally, the method without using the BRF classifier adopts the fully connected classifier (FCC). It designs a fully connected layer with a Softmax function after a well-trained and frozen feature extractor, and the neuron number incrementally increases with the number of faults increasing. We train FCC using cross-entropy loss and Adam optimizer with batches of 512, and a learning rate of 0.001 in 500 epochs. 

As illustrated in \autoref{tab:ablation}, our method achieves the highest average accuracy across all sessions when all three components are utilized for both TEP and MFF datasets. 
Moreover, adding the SCL, MES, and BRF modules individually improves accuracy. For instance, in the TE imbalanced case, SCL increases accuracy from 50.16\% to 80.80\%. MES improves accuracy from 39.39\% to 59.49\% in the TE long-tailed case, while BRF raises it from 50.67\% to 73.93\% in MFF long-tailed case 2, showing each module's impact.
The absence of the SCL module leads to a decrease in accuracy, indicating the significance of SCL in the model's performance.
When the MES module is not used, and Herding, random selection, and Mixed strategy are adopted instead, there are decreases in accuracies, signifying that MES plays a positive role in sample selection. Moreover, the mixed approach may fail to fully capture the benefits of either method, leading to suboptimal performance.
Replacing the BRF classifier with an FCC classifier results in reduced accuracies, suggesting that the BRF classifier contributes positively to the classification performance. In conclusion, experiments demonstrate that all three modules contribute to the performance improvement of SCLIFD.

\subsection{Generalization of Our Method}
The TEP dataset consists of 21 different fault types. To demonstrate the generalizability of our proposed method, we selected an additional 9 fault classes, along with the normal class, from the imbalanced TEP dataset. These were used to form five distinct experimental groups, each comprising different fault classes, to evaluate the classification performance throughout the incremental learning process. As shown in \autoref{tab:class combinations}, the average accuracy across six different class selections from the imbalanced TEP dataset reached 85.86\%. This result demonstrates that our method maintains strong classification accuracy for other fault types beyond the original set, further validating its robustness and adaptability in handling a variety of fault diagnosis scenarios.

\begin{table}[!t]
\scriptsize
\setlength\tabcolsep{4pt}
  \centering
  \caption{Accuracies for Different Faults Types on the Imbalanced TEP.}
  \label{tab:class combinations}
\renewcommand{\arraystretch}{1.3}  
    \begin{tabular}{lcccccc}
    \toprule

    \multirow{2}{*}{Different Fault Types} & \multicolumn{5}{c}{Accuracy(\%) in Incremental Sessions ↑}                                      & \multirow{2}{*}{Average}\\
\cline{2-6}          
           & 1 & 2 & 3 & 4 & 5  &  \\
    \hline
    1/2/6/7/13/17/18/19/20  & 99.20  & 98.45  & 90.42  & 76.97  & 68.06  & 86.62 \\
    1/2/5/6/7/13/17/18/19/20 & 99.15  & 94.91  & 94.46  & 79.36  & 63.60  & 86.30 \\
    1/2/6/7/13/16/17/18/19/20  & 99.32  & 98.72  & 90.87  & 71.99  & 63.25  & 84.83 \\
    1/2/6/7/11/13/17/18/19/20 & 99.15  & 98.72  & 80.75  & 73.73  & 62.35  & 82.94 \\
    1/2/6/7/10/13/17/18/19/20 & 99.09  & 98.15  & 81.88  & 76.22  & 65.89  & 84.26 \\
    \bottomrule
    \end{tabular}%
\end{table}%

\subsection{Sensitivity Evaluation}

\begin{figure}[t]
    \centering
    \includegraphics[width=1\linewidth]{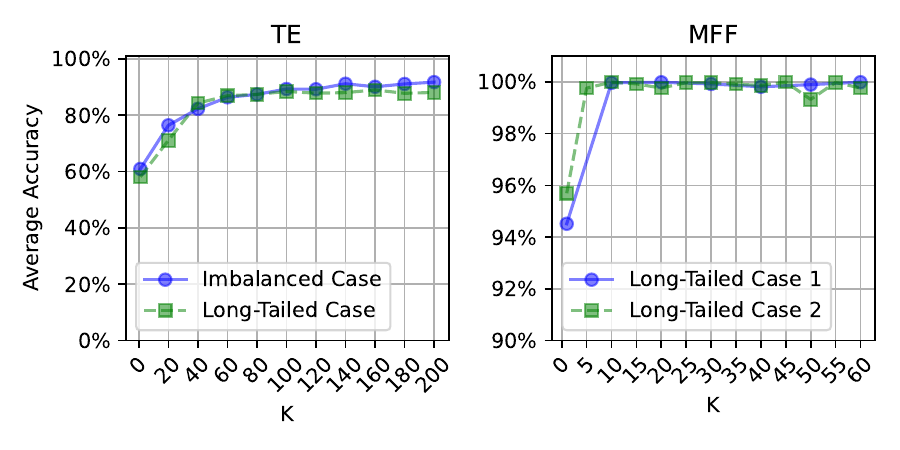}
    \caption{Average Accuracies of different K values under TE and MFF datasets.}
    \label{fig:accuracy_curves_combined}
\end{figure}

To determine the appropriate memory buffer size $K$, we conduct a sensitivity analysis to minimize memory usage while maintaining accuracy. As shown in \autoref{fig:accuracy_curves_combined}, the model's performance remains stable across different $K$ values. For the TE dataset's imbalanced case, accuracy stabilizes between $K=100$ to $200$, so $K=100$ is selected. In the TE long-tailed case, accuracy remains stable from $K=40$ to $200$, leading to $K=40$ being chosen. For the MFF dataset, $K=10$ and $K=5$ are selected for long-tailed cases 1 and 2, as accuracy stabilizes between $K=10$ to $60$ and $K=5$ to $60$, respectively. In each case, the smallest stable $K$ is chosen to optimize memory usage.



\section{Conclusion}
\label{conclusion}
This paper introduces the SCLIFD framework, designed to address fault diagnosis challenges under conditions of limited fault data within a class incremental learning framework. SCLIFD enhances its representation learning capabilities and reduces catastrophic forgetting through the use of supervised contrastive knowledge distillation. Additionally, we introduce a novel sample selection method, MES, which sharpens the model's ability to discern boundaries and enhance generalization. We also incorporate the BRF classifier at the classification stage to counteract the effects of class imbalance.
We assess the SCLIFD in scenarios of imbalanced and long-tailed fault diagnosis, conducting comprehensive experiments on the TEP and MFF datasets. SCLIFD has outperformed other SOTA methods in various scenarios, underscoring its innovative approach to overcoming the difficulties of class incremental learning in fault diagnosis with insufficient data.
In the future, we will explore dynamically adjusting the memory buffer size for each class based on data characteristics and model needs. This could help optimize memory usage and maintain performance as the number of data categories increases.


%


\ifCLASSOPTIONcaptionsoff
  \newpage
\fi



%
\footnotesize{
\bibliographystyle{IEEEtranN}

\bibliography{ref.bib}

\begin{thebibliography}{46}
\providecommand{\natexlab}[1]{#1}
\providecommand{\url}[1]{#1}
\csname url@samestyle\endcsname
\providecommand{\newblock}{\relax}
\providecommand{\bibinfo}[2]{#2}
\providecommand{\BIBentrySTDinterwordspacing}{\spaceskip=0pt\relax}
\providecommand{\BIBentryALTinterwordstretchfactor}{4}
\providecommand{\BIBentryALTinterwordspacing}{\spaceskip=\fontdimen2\font plus
\BIBentryALTinterwordstretchfactor\fontdimen3\font minus \fontdimen4\font\relax}
\providecommand{\BIBforeignlanguage}[2]{{%
\expandafter\ifx\csname l@#1\endcsname\relax
\typeout{** WARNING: IEEEtranN.bst: No hyphenation pattern has been}%
\typeout{** loaded for the language `#1'. Using the pattern for}%
\typeout{** the default language instead.}%
\else
\language=\csname l@#1\endcsname
\fi
#2}}
\providecommand{\BIBdecl}{\relax}
\BIBdecl

\bibitem[An et~al.(2024)An, Zhang, Chai, Zhu, and Liu]{10106026}
Y.~An, K.~Zhang, Y.~Chai, Z.~Zhu, and Q.~Liu, ``Gaussian mixture variational-based transformer domain adaptation fault diagnosis method and its application in bearing fault diagnosis,'' \emph{IEEE Transactions on Industrial Informatics}, vol.~20, no.~1, pp. 615--625, 2024.

\bibitem[Gao et~al.(2022)Gao, Gao, Li, and Cao]{9780560}
Y.~Gao, L.~Gao, X.~Li, and S.~Cao, ``A hierarchical training-convolutional neural network for imbalanced fault diagnosis in complex equipment,'' \emph{IEEE Transactions on Industrial Informatics}, vol.~18, no.~11, pp. 8138--8145, 2022.

\bibitem[Li et~al.(2023{\natexlab{a}})Li, Peng, Sun, Wang, and Wang]{li2023order}
M.~Li, P.~Peng, H.~Sun, M.~Wang, and H.~Wang, ``An order-invariant and interpretable dilated convolution neural network for chemical process fault detection and diagnosis,'' \emph{IEEE Transactions on Automation Science and Engineering}, 2023.

\bibitem[Wang et~al.(2023{\natexlab{a}})Wang, Qin, Li, Butala, Wang, Peng, and Wang]{wang2023hard}
Z.~Wang, B.~Qin, M.~Li, M.~D. Butala, H.~Wang, P.~Peng, and H.~Wang, ``Hard sample mining enabled contrastive feature learning for wind turbine pitch system fault diagnosis,'' \emph{arXiv preprint arXiv:2306.14701}, 2023.

\bibitem[Wang et~al.(2024)Wang, Zhang, Qiao, Ma, Tao, Peng, and Wang]{10510599}
X.~Wang, H.~Zhang, X.~Qiao, K.~Ma, S.~Tao, P.~Peng, and H.~Wang, ``Generalized out-of-distribution fault diagnosis (goofd) via internal contrastive learning,'' \emph{IEEE Transactions on Industrial Informatics}, vol.~20, no.~8, pp. 9987--9996, 2024.

\bibitem[Peng et~al.(2023{\natexlab{a}})Peng, Zhang, Wang, Huang, and Wang]{10114639}
P.~Peng, H.~Zhang, X.~Wang, W.~Huang, and H.~Wang, ``Imbalanced chemical process fault diagnosis using balancing gan with active sample selection,'' \emph{IEEE Sensors Journal}, vol.~23, no.~13, pp. 14\,826--14\,833, 2023.

\bibitem[Huang et~al.(2023)Huang, Zhang, Peng, and Wang]{10152774}
W.~Huang, H.~Zhang, P.~Peng, and H.~Wang, ``Multi-gate mixture-of-expert combined with synthetic minority over-sampling technique for multimode imbalanced fault diagnosis,'' in \emph{2023 26th International Conference on Computer Supported Cooperative Work in Design (CSCWD)}, 2023, pp. 456--461.

\bibitem[Li et~al.(2023{\natexlab{b}})Li, Peng, Zhang, Wang, and Shen]{li2023sccam}
M.~Li, P.~Peng, J.~Zhang, H.~Wang, and W.~Shen, ``Sccam: Supervised contrastive convolutional attention mechanism for ante-hoc interpretable fault diagnosis with limited fault samples,'' \emph{IEEE Transactions on Neural Networks and Learning Systems}, 2023.

\bibitem[Wang et~al.(2023{\natexlab{b}})Wang, Qin, Sun, Zhang, Butala, Demartino, Peng, and Wang]{wang2023imbalanced}
Z.~Wang, B.~Qin, H.~Sun, J.~Zhang, M.~D. Butala, C.~Demartino, P.~Peng, and H.~Wang, ``An imbalanced semi-supervised wind turbine blade icing detection method based on contrastive learning,'' \emph{Renewable Energy}, vol. 212, pp. 251--262, 2023.

\bibitem[Chen et~al.(2022)Chen, Chen, Feng, Liu, Zhang, Zhang, and Xiao]{Chen_Chen_Feng_Liu_Zhang_Zhang_Xiao_2022}
Z.~Chen, J.~Chen, Y.~Feng, S.~Liu, T.~Zhang, K.~Zhang, and W.~Xiao, ``\BIBforeignlanguage{en}{Imbalance fault diagnosis under long-tailed distribution: Challenges, solutions and prospects},'' \emph{\BIBforeignlanguage{en}{Knowledge-Based Systems}}, vol. 258, Dec 2022.

\bibitem[Khoshgoftaar and Gao(2009)]{Khoshgoftaar_Gao_2009a}
T.~M. Khoshgoftaar and K.~Gao, ``Feature selection with imbalanced data for software defect prediction,'' in \emph{2009 International Conference on Machine Learning and Applications}, Dec 2009, pp. 235--240.

\bibitem[Liu et~al.(2017)Liu, Li, and Zio]{Liu_Li_Zio_2017}
J.~Liu, Y.-F. Li, and E.~Zio, ``\BIBforeignlanguage{en}{A svm framework for fault detection of the braking system in a high speed train},'' \emph{\BIBforeignlanguage{en}{Mechanical Systems and Signal Processing}}, vol.~87, pp. 401--409, Mar 2017.

\bibitem[Goodfellow et~al.(2020)Goodfellow, Pouget-Abadie, Mirza, Xu, Warde-Farley, Ozair, Courville, and Bengio]{Goodfellow_Pouget-Abadie_Mirza_Xu_Warde-Farley_Ozair_Courville_Bengio_2020}
I.~Goodfellow, J.~Pouget-Abadie, M.~Mirza, B.~Xu, D.~Warde-Farley, S.~Ozair, A.~Courville, and Y.~Bengio, ``Generative adversarial networks,'' \emph{Communications of the ACM}, vol.~63, no.~11, p. 139–144, Oct 2020.

\bibitem[Li et~al.(2021)Li, Jiang, Liu, Zhang, and Xu]{Li_Jiang_Liu_Zhang_Xu_2021}
X.~Li, H.~Jiang, S.~Liu, J.~Zhang, and J.~Xu, ``\BIBforeignlanguage{en}{A unified framework incorporating predictive generative denoising autoencoder and deep coral network for rolling bearing fault diagnosis with unbalanced data},'' \emph{\BIBforeignlanguage{en}{Measurement}}, vol. 178, Jun 2021.

\bibitem[Zhiyi et~al.(2020)Zhiyi, Haidong, Lin, Junsheng, and Yu]{Zhiyi_Haidong_Lin_Junsheng_Yu_2020}
H.~Zhiyi, S.~Haidong, J.~Lin, C.~Junsheng, and Y.~Yu, ``\BIBforeignlanguage{en}{Transfer fault diagnosis of bearing installed in different machines using enhanced deep auto-encoder},'' \emph{\BIBforeignlanguage{en}{Measurement}}, vol. 152, Feb 2020.

\bibitem[Chu et~al.(2020)Chu, Bian, Liu, and Ling]{chu2020feature}
P.~Chu, X.~Bian, S.~Liu, and H.~Ling, ``Feature space augmentation for long-tailed data,'' in \emph{European Conference on Computer Vision}.\hskip 1em plus 0.5em minus 0.4em\relax Springer, 2020, pp. 694--710.

\bibitem[McCloskey and Cohen(1989)]{McCloskey_Cohen_1989}
M.~McCloskey and N.~J. Cohen, \emph{\BIBforeignlanguage{en}{Catastrophic Interference in Connectionist Networks: The Sequential Learning Problem}}.\hskip 1em plus 0.5em minus 0.4em\relax Academic Press, Jan 1989, vol.~24, pp. 109--165.

\bibitem[Yao and Zhang(2024)]{yao2024uncertaintyclarityuncertaintyguidedclassincremental}
\BIBentryALTinterwordspacing
Y.~Yao and H.~Zhang, ``From uncertainty to clarity: Uncertainty-guided class-incremental learning for limited biomedical samples via semantic expansion,'' 2024. [Online]. Available: \url{https://arxiv.org/abs/2409.07757}
\BIBentrySTDinterwordspacing

\bibitem[Peng et~al.(2023{\natexlab{b}})Peng, Zhang, Li, Peng, Wang, and Shen]{peng2023sclifdsupervisedcontrastiveknowledgedistillation}
\BIBentryALTinterwordspacing
P.~Peng, H.~Zhang, M.~Li, G.~Peng, H.~Wang, and W.~Shen, ``Sclifd:supervised contrastive knowledge distillation for incremental fault diagnosis under limited fault data,'' 2023. [Online]. Available: \url{https://arxiv.org/abs/2302.05929}
\BIBentrySTDinterwordspacing

\bibitem[Shi et~al.(2024)Shi, Ding, Chang, Shen, Huang, and Zhu]{Shi_Ding_Chang_Shen_Huang_Zhu_2024}
M.~Shi, C.~Ding, S.~Chang, C.~Shen, W.~Huang, and Z.~Zhu, ``Cross-domain class incremental broad network for continuous diagnosis of rotating machinery faults under variable operating conditions,'' \emph{IEEE Transactions on Industrial Informatics}, vol.~20, no.~4, pp. 6356--6368, Apr. 2024.

\bibitem[Hell et~al.(2022)Hell, Pestana~de Aguiar, Soares, and Goliatt]{Hell_Pestana_Soares_Goliatt_2022}
M.~Hell, E.~Pestana~de Aguiar, N.~Soares, and L.~Goliatt, ``\BIBforeignlanguage{en}{A data-driven time-series fault prediction framework for dynamically evolving large-scale data streaming systems},'' \emph{\BIBforeignlanguage{en}{International Journal of Fuzzy Systems}}, vol.~24, no.~6, pp. 2831--2844, Sep 2022.

\bibitem[Zheng et~al.(2022)Zheng, Xiong, Zhang, Su, and Hu]{Zheng_Xiong_Zhang_Su_Hu_2022}
J.~Zheng, H.~Xiong, Y.~Zhang, K.~Su, and Z.~Hu, ``\BIBforeignlanguage{en}{Bearing fault diagnosis via incremental learning based on the repeated replay using memory indexing (r-remind) method},'' \emph{\BIBforeignlanguage{en}{Machines}}, vol.~10, no.~55, May 2022.

\bibitem[Yu and Zhao(2020)]{Yu_Zhao_2020}
W.~Yu and C.~Zhao, ``Broad convolutional neural network based industrial process fault diagnosis with incremental learning capability,'' \emph{IEEE Transactions on Industrial Electronics}, vol.~67, no.~6, pp. 5081--5091, Jun 2020.

\bibitem[Gu et~al.(2022)Gu, Zhao, Yang, and Li]{Gu_Zhao_Yang_Li_2022}
X.~Gu, Y.~Zhao, G.~Yang, and L.~Li, ``\BIBforeignlanguage{en}{An imbalance modified convolutional neural network with incremental learning for chemical fault diagnosis},'' \emph{\BIBforeignlanguage{en}{IEEE Transactions on Industrial Informatics}}, vol.~18, no.~6, pp. 3630--3639, Jun. 2022.

\bibitem[Ren et~al.(2022)Ren, Liu, Wang, and Zhang]{Ren_Liu_Wang_Zhang_2022}
Y.~Ren, J.~Liu, Q.~Wang, and H.~Zhang, ``Hsell-net: A heterogeneous sample enhancement network with lifelong learning under industrial small samples,'' \emph{IEEE Transactions on Cybernetics}, pp. 1--13, 2022.

\bibitem[Shao et~al.(2021)Shao, Xia, Han, Zhang, and Wan]{shaoIFDCNN}
H.~Shao, M.~Xia, G.~Han, Y.~Zhang, and J.~Wan, ``Intelligent fault diagnosis of rotor-bearing system under varying working conditions with modified transfer convolutional neural network and thermal images,'' \emph{IEEE Transactions on Industrial Informatics}, vol.~17, no.~5, pp. 3488--3496, 2021.

\bibitem[Zhang et~al.(2021)Zhang, Li, Ma, Luo, and Li]{zhangOpenDA}
W.~Zhang, X.~Li, H.~Ma, Z.~Luo, and X.~Li, ``Open-set domain adaptation in machinery fault diagnostics using instance-level weighted adversarial learning,'' \emph{IEEE Transactions on Industrial Informatics}, vol.~17, no.~11, pp. 7445--7455, 2021.

\bibitem[Liu et~al.(2022)Liu, Wang, Chow, and Li]{liuDeepSunDA}
Y.~Liu, Y.~Wang, T.~W.~S. Chow, and B.~Li, ``Deep adversarial subdomain adaptation network for intelligent fault diagnosis,'' \emph{IEEE Transactions on Industrial Informatics}, vol.~18, no.~9, pp. 6038--6046, 2022.

\bibitem[Liu et~al.(2024)Liu, Zheng, and Liang]{liangVariable}
L.~Liu, Y.~Zheng, and S.~Liang, ``Variable-wise stacked temporal autoencoder for intelligent fault diagnosis of industrial systems,'' \emph{IEEE Transactions on Industrial Informatics}, vol.~20, no.~5, pp. 7545--7555, 2024.

\bibitem[Zhang et~al.(2024)Zhang, Liu, Zhang, and Lu]{Zhang2024MultiscaleCA}
X.~Zhang, J.~Liu, X.~Zhang, and Y.~Lu, ``Multiscale channel attention-driven graph dynamic fusion learning method for robust fault diagnosis,'' \emph{IEEE Transactions on Industrial Informatics}, vol.~20, pp. 11\,002--11\,013, 2024.

\bibitem[Khosla et~al.(2020)Khosla, Teterwak, Wang, Sarna, Tian, Isola, Maschinot, Liu, and Krishnan]{Khosla_Teterwak_Wang_Sarna_Tian_Isola_Maschinot_Liu_Krishnan_2020}
P.~Khosla, P.~Teterwak, C.~Wang, A.~Sarna, Y.~Tian, P.~Isola, A.~Maschinot, C.~Liu, and D.~Krishnan, ``Supervised contrastive learning,'' in \emph{Advances in Neural Information Processing Systems}, vol.~33.\hskip 1em plus 0.5em minus 0.4em\relax Curran Associates, Inc., 2020, pp. 18\,661--18\,673.

\bibitem[Chen et~al.(2004)Chen, Liaw, and Breiman]{Chen2004UsingRF}
C.~Chen, A.~Liaw, and L.~Breiman, ``Using random forest to learn imbalanced data,'' Univ. California, Berkeley, Berkeley, CA, USA, Tech. Rep., 2004.

\bibitem[Rebuffi et~al.(2017)Rebuffi, Kolesnikov, Sperl, and Lampert]{icarl}
S.-A. Rebuffi, A.~Kolesnikov, G.~Sperl, and C.~H. Lampert, ``i\text{C}a\text{RL}: Incremental classifier and representation learning,'' in \emph{2017 IEEE Conference on Computer Vision and Pattern Recognition (CVPR)}, 2017, pp. 5533--5542.

\bibitem[Welling(2009)]{welling2009herding}
M.~Welling, ``Herding dynamical weights to learn,'' in \emph{Proceedings of the 26th Annual International Conference on Machine Learning}, 2009, pp. 1121--1128.

\bibitem[Castro et~al.(2018)Castro, Marin-Jimenez, Guil, Schmid, and Alahari]{Castro_2018_ECCV}
F.~M. Castro, M.~J. Marin-Jimenez, N.~Guil, C.~Schmid, and K.~Alahari, ``End-to-end incremental learning,'' in \emph{Proceedings of the European Conference on Computer Vision (ECCV)}, Sept 2018.

\bibitem[Li et~al.(2024)Li, Peng, Zhang, Wang, and Shen]{Li_Peng_Zhang_Wang_Shen_2024}
M.~Li, P.~Peng, J.~Zhang, H.~Wang, and W.~Shen, ``\BIBforeignlanguage{en}{Sccam: Supervised contrastive convolutional attention mechanism for ante-hoc interpretable fault diagnosis with limited fault samples},'' \emph{\BIBforeignlanguage{en}{IEEE Transactions on Neural Networks and Learning Systems}}, pp. 1--12, 2024.

\bibitem[Breiman(2017)]{breiman2017classification}
L.~Breiman, \emph{Classification and regression trees}.\hskip 1em plus 0.5em minus 0.4em\relax Routledge, 2017.

\bibitem[Wu et~al.(2019)Wu, Chen, Wang, Ye, Liu, Guo, and Fu]{Wu_2019_CVPR}
Y.~Wu, Y.~Chen, L.~Wang, Y.~Ye, Z.~Liu, Y.~Guo, and Y.~Fu, ``Large scale incremental learning,'' in \emph{Proceedings of the IEEE/CVF Conference on Computer Vision and Pattern Recognition (CVPR)}, June 2019.

\bibitem[Peng et~al.(2022)Peng, Lu, Tao, Ma, Zhang, Wang, and Zhang]{Peng_Lu_Tao_Ma_Zhang_Wang_Zhang_2022}
P.~Peng, J.~Lu, S.~Tao, K.~Ma, Y.~Zhang, H.~Wang, and H.~Zhang, ``Progressively balanced supervised contrastive representation learning for long-tailed fault diagnosis,'' \emph{IEEE Transactions on Instrumentation and Measurement}, vol.~71, pp. 1--12, 2022.

\bibitem[Gou et~al.(2021)Gou, Yu, Maybank, and Tao]{Gou_Yu_Maybank_Tao_2021}
J.~Gou, B.~Yu, S.~J. Maybank, and D.~Tao, ``\BIBforeignlanguage{en}{Knowledge distillation: A survey},'' \emph{\BIBforeignlanguage{en}{International Journal of Computer Vision}}, vol. 129, no.~6, pp. 1789--1819, Jun. 2021.

\bibitem[{Lawrence Ricker}(1996)]{LAWRENCERICKER1996205}
N.~{Lawrence Ricker}, ``Decentralized control of the tennessee eastman challenge process,'' \emph{Journal of Process Control}, vol.~6, no.~4, pp. 205--221, 1996.

\bibitem[Ruiz-C{\'a}rcel et~al.(2015)Ruiz-C{\'a}rcel, Cao, Mba, Lao, and Samuel]{ruiz2015statistical}
C.~Ruiz-C{\'a}rcel, Y.~Cao, D.~Mba, L.~Lao, and R.~Samuel, ``Statistical process monitoring of a multiphase flow facility,'' \emph{Control Engineering Practice}, vol.~42, pp. 74--88, 2015.

\bibitem[Li and Hoiem(2018)]{Li2018}
Z.~Li and D.~Hoiem, ``Learning without forgetting,'' \emph{IEEE Transactions on Pattern Analysis and Machine Intelligence}, vol.~40, pp. 2935--2947, Dec 2018.

\bibitem[Song et~al.(2023)Song, Zhao, Shi, Peng, Yuan, and Tian]{song2023learning}
Z.~Song, Y.~Zhao, Y.~Shi, P.~Peng, L.~Yuan, and Y.~Tian, ``Learning with fantasy: Semantic-aware virtual contrastive constraint for few-shot class-incremental learning,'' in \emph{Proceedings of the IEEE/CVF conference on computer vision and pattern recognition}, 2023, pp. 24\,183--24\,192.

\bibitem[Kim et~al.(2023)Kim, Han, Seo, and Moon]{kim2023warping}
D.-Y. Kim, D.-J. Han, J.~Seo, and J.~Moon, ``Warping the space: Weight space rotation for class-incremental few-shot learning,'' in \emph{The Eleventh International Conference on Learning Representations}, 2023.

\bibitem[Zhao et~al.(2023)Zhao, Lu, Xu, Cheng, Guo, Niu, and Fang]{zhao2023few}
L.~Zhao, J.~Lu, Y.~Xu, Z.~Cheng, D.~Guo, Y.~Niu, and X.~Fang, ``Few-shot class-incremental learning via class-aware bilateral distillation,'' in \emph{Proceedings of the IEEE/CVF conference on computer vision and pattern recognition}, 2023, pp. 11\,838--11\,847.

\end{thebibliography}
}
\vspace{12pt}

\begin{IEEEbiography}[{\includegraphics[width=1in,height=1.25in,clip,keepaspectratio]{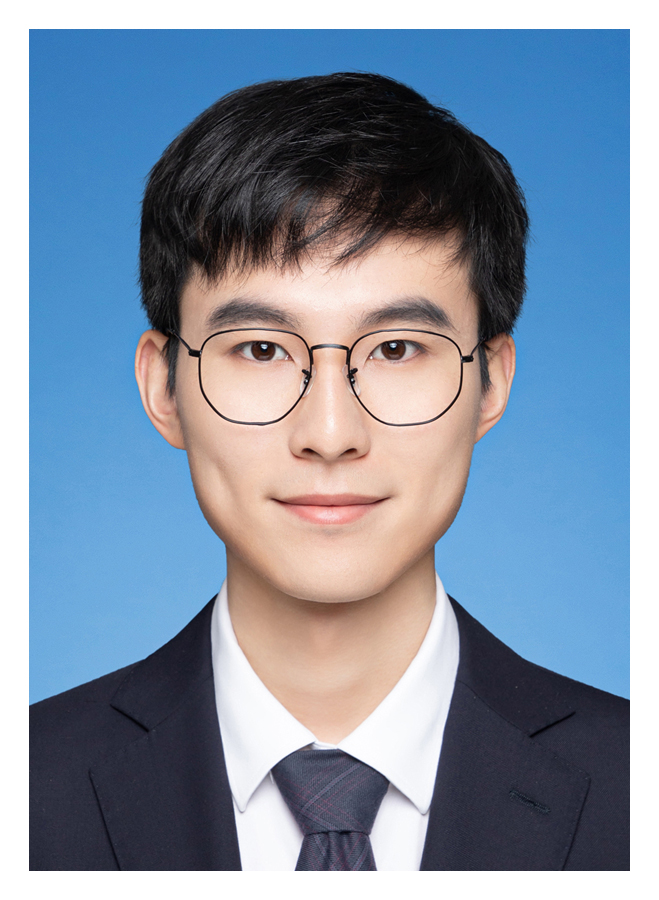}}]{Hanrong Zhang} (\url{https://zhang-henry.github.io/}) received a dual B.S. degree in Computer Science from the University of Leeds, United Kingdom, and Southwest Jiaotong University, China, in 2022. He is currently pursuing a Master's degree in Computer Engineering at the ZJU-UIUC Joint Institute, Zhejiang University, Haining, China. His research interests include fault diagnosis and trustworthy AI. 
\end{IEEEbiography}

\begin{IEEEbiography}[{\includegraphics[width=1in,height=1.25in,clip,keepaspectratio]{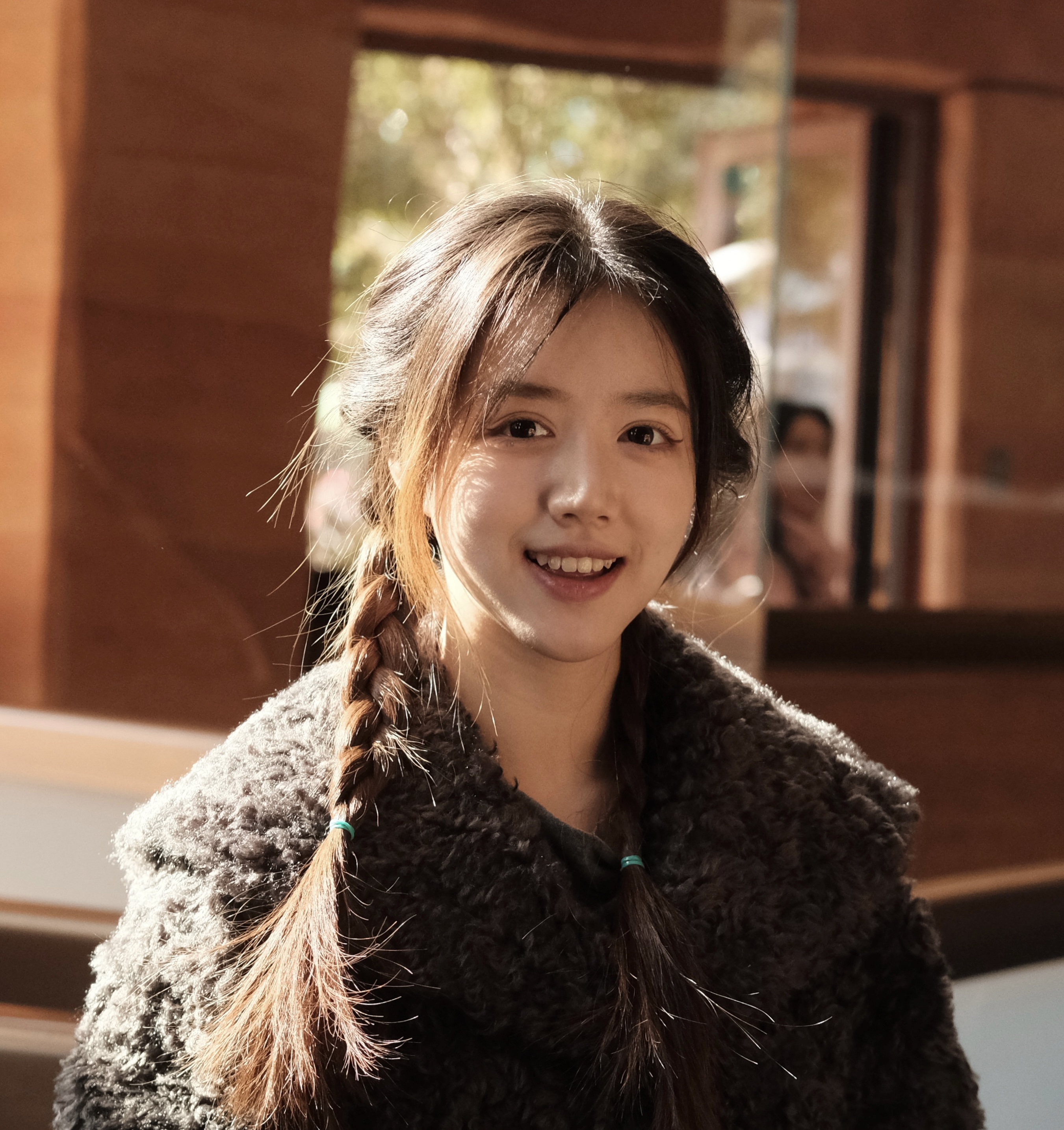}}]{Yifei Yao} received a dual B.S. degree in Computer Science from the University of Leeds, United Kingdom, and Southwest Jiaotong University, China, in 2022. She is currently pursuing a Master's degree at Zhejiang University, China. Her research interest is deep learning. 
\end{IEEEbiography}

\begin{IEEEbiography}[{\includegraphics[width=1in,height=1.25in,clip,keepaspectratio]{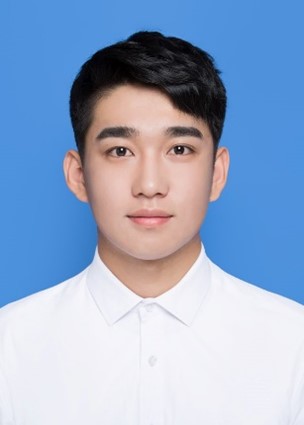}}]{Zixuan Wang} is currently a doctoral student at the College of Biomedical Engineering and Instrument Science in Zhejiang University, Hangzhou, China. He received his Bachelor's degree at Tsien Hsue-shen College from Nanjing University of Science and Technology in 2020. His research interests are deep learning and data-driven fault diagnosis.
\end{IEEEbiography}

\begin{IEEEbiography}[{\includegraphics[width=1in,height=1.25in,clip,keepaspectratio]{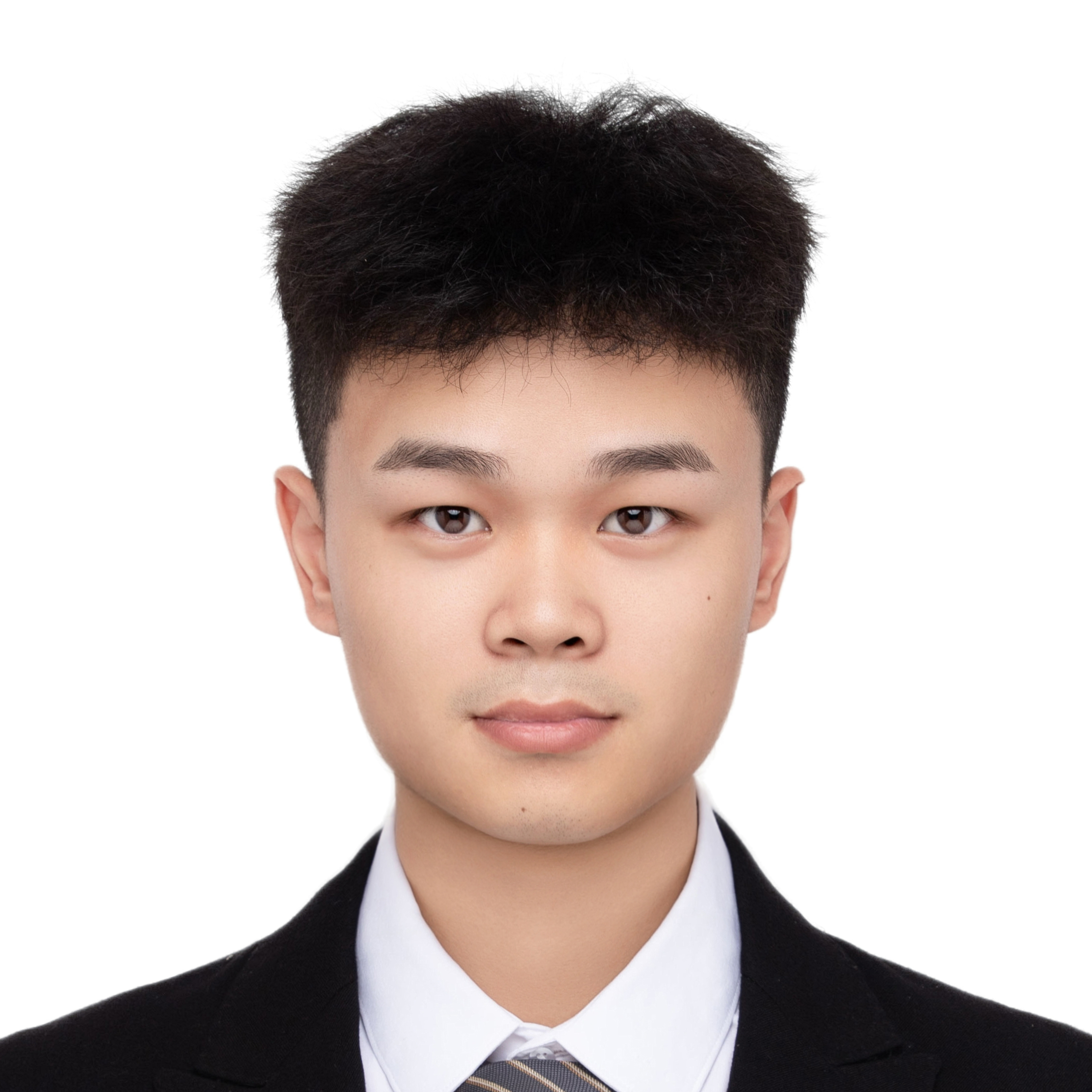}}]{Jiayuan Su} is currently pursuing his master's degree in Artificial Intelligence at the ZJU-UIUC Institute, Zhejiang University. Previously, he earned a B.Sc. degree in Robotics and Intelligent Devices from the National University of Ireland Maynooth in 2023. His research interests are primarily focused on deep learning and natural language processing.
\end{IEEEbiography}

\begin{IEEEbiography}[{\includegraphics[width=1in,height=1.25in,clip,keepaspectratio]{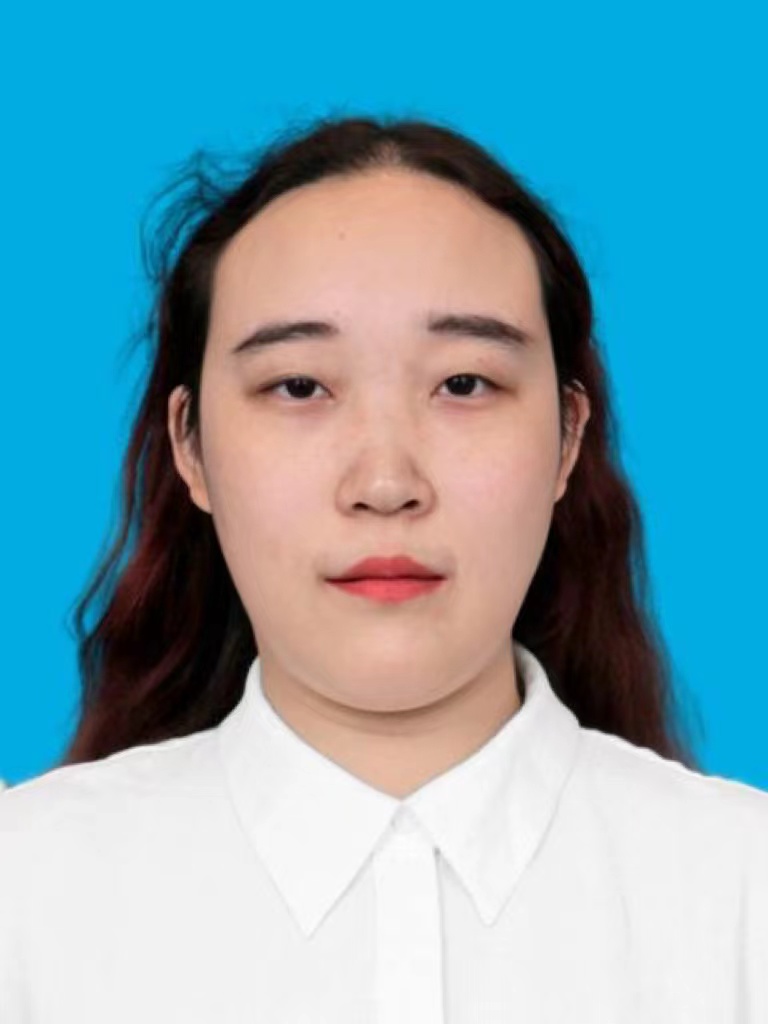}}]{Mengxuan Li} received a dual bachelor’s degree in computer engineering from Zhejiang University, Hangzhou, China, and the University of Illinois at Urbana-Champaign, Champaign, IL, USA, in 2020. She is pursuing a Doctoral degree with the College of Computer Science and Technology, Zhejiang University. Her research interests include explainable artificial intelligence (XAI), machine learning (ML), and long-tail learning.
\end{IEEEbiography}

\begin{IEEEbiography}[{\includegraphics[width=1in,height=1.25in,clip,keepaspectratio]{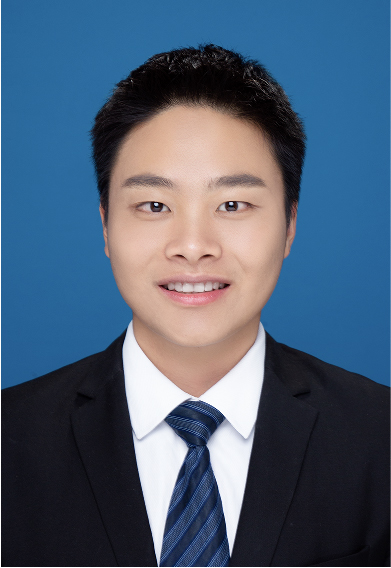}}]{Peng Peng} is currently a doctoral student at the National Engineering Research Centre of Computer Integrated Manufacturing System (CIMS-ERC) in Tsinghua University, Beijing, China.  He received his Bachelor's degree in the Department of Automation from Northeastern University in 2016. His research interests are process monitoring and prognostic and health management. 
\end{IEEEbiography}

\begin{IEEEbiography}[{\includegraphics[width=1in,height=1.25in, clip,keepaspectratio]{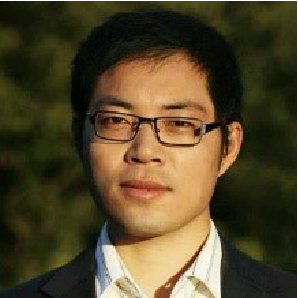}}]{Hongwei Wang} received the B.S. degree in information technology and instrumentation from Zhejiang University, China, in 2004, the M.S. degree in control science and engineering from Tsinghua University, China, in 2007, and the Ph.D. degree in design knowledge retrieval from the University of Cambridge. From 2011 to 2018, he was a Lecturer and then, a Senior Lecturer in engineering design at the University of Portsmouth. He is currently a Tenured Professor at Zhejiang University and the University of Illinois at Urbana–Champaign Joint Institute. His research interests include knowledge engineering, industrial knowledge graphs, intelligent and collaborative systems, and data-driven fault diagnosis. His research in these areas has been published in over 120 peer-reviewed papers in well-established journals and international conferences. 
\end{IEEEbiography}




\end{document}